\documentclass[english]{article}
\usepackage{geometry}
\usepackage{color}
\usepackage{xcolor}
\usepackage{url}
\usepackage{babel}
\usepackage{booktabs}
\usepackage{multirow}
\usepackage{makecell}
\usepackage{siunitx}
\usepackage{mathtools}
\usepackage{bm}
\usepackage{bbm}
\usepackage{amsmath}
\usepackage{amsthm}
\usepackage{amssymb}
\usepackage{algorithm}
\usepackage{algorithmic}
\usepackage{graphicx}
\usepackage{caption}
\usepackage{subcaption}
\usepackage{appendix}
\usepackage{xspace}
\usepackage{verbatim}
\usepackage{wrapfig}
\usepackage{tabularx}

\usepackage[unicode=true,pdfusetitle,bookmarks=true,bookmarksnumbered=true,bookmarksopen=true,bookmarksopenlevel=1,
 breaklinks=false,pdfborder={0 0 0},pdfborderstyle={},backref=false,colorlinks=true]{hyperref}
\hypersetup{
 linkcolor=blue, urlcolor=blue, citecolor=blue, linktoc=all, linktocpage=false}
\hypersetup{bookmarksopen=true,
        bookmarksopenlevel=1, 
        bookmarksnumbered=true, }

\usepackage[align=center,shadow=false,shadowsize=5pt,nobreak=true,framemethod=tikz,style=0,skipabove=4pt,skipbelow=1pt,innertopmargin=-1pt,innerbottommargin=4pt,innerleftmargin=5pt,innerrightmargin=5pt,leftmargin=0pt,rightmargin=0pt]{mdframed}
\usepackage{rotating} %

\usepackage{placeins}

\usepackage{natbib}
 
\usepackage{listings}
\makeatletter

\theoremstyle{plain}

\theoremstyle{remark}
\newtheorem{remark}{\protect\remarkname}
\theoremstyle{plain}

\theoremstyle{plain}

\theoremstyle{plain}

\theoremstyle{plain}

\theoremstyle{plain}

\theoremstyle{definition}

\theoremstyle{definition}

\theoremstyle{plain}

\definecolor{todo}{RGB}{200,0,20}

\definecolor{new}{RGB}{200,0,200}

\definecolor{emerald}{rgb}{0.31, 0.78, 0.47}

\definecolor{yanxi}{RGB}{100,0,200}

\let\bar\overline

\allowdisplaybreaks

\renewcommand{\Pr}{{\mathbb{P}}}

\newcommand{\ind}{\mathbbm{1}}

\newcommand{\bg}{\bm{g}}

\newcommand{\bz}{\bm{z}}

\newcommand{\btheta}{\bm{\theta}}

\newcommand{\trinity}{\texttt{Trinity-RFT}\xspace}

\newcommand{\cod}{\textsc{CoD}\xspace}
\newcommand{\codtrain}{\textsc{CoD-Train}\xspace}
\newcommand{\coddeploy}{\textsc{CoD-Deploy}\xspace}

\newcommand{\frozenlake}{\textsc{FrozenLake}\xspace}
\newcommand{\flobs}{\textsc{FrozenLake-Obscure}\xspace}
\newcommand{\alchemy}{\textsc{Alchemy}\xspace}
\newcommand{\alchrand}{\textsc{Alchemy-Random}\xspace}
\newcommand{\terminal}{\textsc{TerminalSimulator}\xspace}
\newcommand{\pde}{\textsc{PDE-Discovery}\xspace}
\newcommand{\control}{\textsc{OptimalControl}\xspace}
\newcommand{\navigation}{\textsc{GridNavigation}\xspace}

\newcommand{\qweneight}{Qwen3-8B\xspace}
\newcommand{\qwensix}{Qwen3.6-27B\xspace}
\newcommand{\qwentpemax}{Qwen3.8-Max\xspace}
\newcommand{\qwentpeseven}{Qwen3.8-27B\xspace}

\newcommand*\diff{\mathop{}\!\mathrm{d}}

\newcommand{\pitheta}{\pi_{\btheta}}

\newcommand{\piold}{\pi_{\textsf{old}}}

\newcommand{\epslow}{\epsilon_{\textsf{low}}}
\newcommand{\epshigh}{\epsilon_{\textsf{high}}}

\newcommand{\nablatheta}{\nabla_{\btheta}}

\newcommand{\rec}{\textsc{REC}\xspace}
\newcommand{\reconesidenois}{\textsc{REC-OneSide-NoIS}\xspace}

\newcommand{\redweight}{\textsc{RED-Weight}\xspace}

\newcommand{\rx}{r^{x}}
\newcommand{\rz}{r^{z}}
\newcommand{\Rx}{R^{x}}
\newcommand{\Rz}{R^{z}}
\newcommand{\Rbarx}{\bar{R}^{x}}
\newcommand{\Rbarz}{\bar{R}^{z}}
\newcommand{\Ax}{A^{x}}
\newcommand{\Az}{A^{z}}

\newcommand{\Ai}{A_i}
\newcommand{\pioldi}{\piold(i)}
\newcommand{\pithetai}{\pitheta(i)}
\newcommand{\Mi}{M_i}

\makeatother

\providecommand{\assumptionname}{Assumption}
\providecommand{\corollaryname}{Corollary}
\providecommand{\examplename}{Example}
\providecommand{\definitionname}{Definition}
\providecommand{\factname}{Fact}
\providecommand{\lemmaname}{Lemma}
\providecommand{\propositionname}{Proposition}
\providecommand{\remarkname}{Remark}
\providecommand{\theoremname}{Theorem}

\begin{document}
\title{Connect the Dots: Training LLMs for Long-Lifecycle Agents with Cross-Domain Generalization Via Reinforcement Learning}

\author{
  Yanxi Chen$^*$$^{1}$,
  Weijie Shi$^*$$^{2}$,
  Boyi Hu$^{3}$, 
  Zeyue Zhang$^{4}$, 
  Zhiwei Wang$^{5}$, \\
  Yuexiang Xie$^{1}$,
  Yaliang Li$^{1}$,
  Bolin Ding$^{1}$,
  Jingren Zhou$^{1}$ \\
  \ \\
  Alibaba Group
}
\date{}

\maketitle

\renewcommand*{\thefootnote}{\fnsymbol{footnote}}
\footnotetext[1]{Equal contribution.}
\renewcommand*{\thefootnote}{\arabic{footnote}}
\footnotetext[1]{Alibaba Group. Contact: \texttt{yxchen0@outlook.com}, \texttt{\{yaliang.li, bolin.ding\}@alibaba-inc.com}}
\footnotetext[2]{Hong Kong University of Science and Technology. Work done during an internship at Alibaba Group.}
\footnotetext[3]{The University of Hong Kong. Work done during an internship at Alibaba Group.}
\footnotetext[4]{Renmin University of China. Work done during an internship at Alibaba Group.}
\footnotetext[5]{Tsinghua University. Work done during an internship at Alibaba Group.}

\vspace{-1.5em}

\begin{abstract}

This work presents a general framework for training large language models (LLMs) to ``\underline{Co}nnect the \underline{D}ots'' (\cod), a meta-capability required by long-lifecycle agents:
as an LLM-based AI agent gets deployed in an environment, it solves a long sequence of tasks while continuously exploring the environment, learning from its own experiences, and iteratively self-updating its context about the environment, thereby achieving progressively better performance on future tasks conditioned on the updated context.
Major components of the \cod framework include:
(1) algorithm design and infrastructure for end-to-end reinforcement learning (RL) with long rollout sequences interleaving solve-task and update-context episodes;
(2) tasks and environments for incentivizing and eliciting the targeted meta-capability in LLMs during training, as well as for faithfully measuring progress during evaluation.
We present proof-of-concept implementations of the \cod framework,
including a GRPO-style RL algorithm with fine-grained credit assignment,
as well as tasks and environments tailored to the targeted meta-capability (rather than domain-specific LLM capabilities or standard task-by-task RL).
Empirical results with \qweneight and \qwensix validate the efficacy of end-to-end RL training in the \cod setting,
and demonstrate the potential for out-of-distribution generalization ---
within the training domains, across different domains, and from \cod to Ralph-loop settings
--- of the elicited meta-capability.
Our investigation of \cod connects several lines of prior work,
and opens up new opportunities for advancing LLMs and AI agents.
To facilitate further research and applications,
we release our implementations at \url{https://github.com/agentscope-ai/Trinity-RFT/tree/research/cod_v2/examples/research_cod}.

\end{abstract}

\begin{figure}[H]
    \centering

    \includegraphics[width=.31\textwidth]{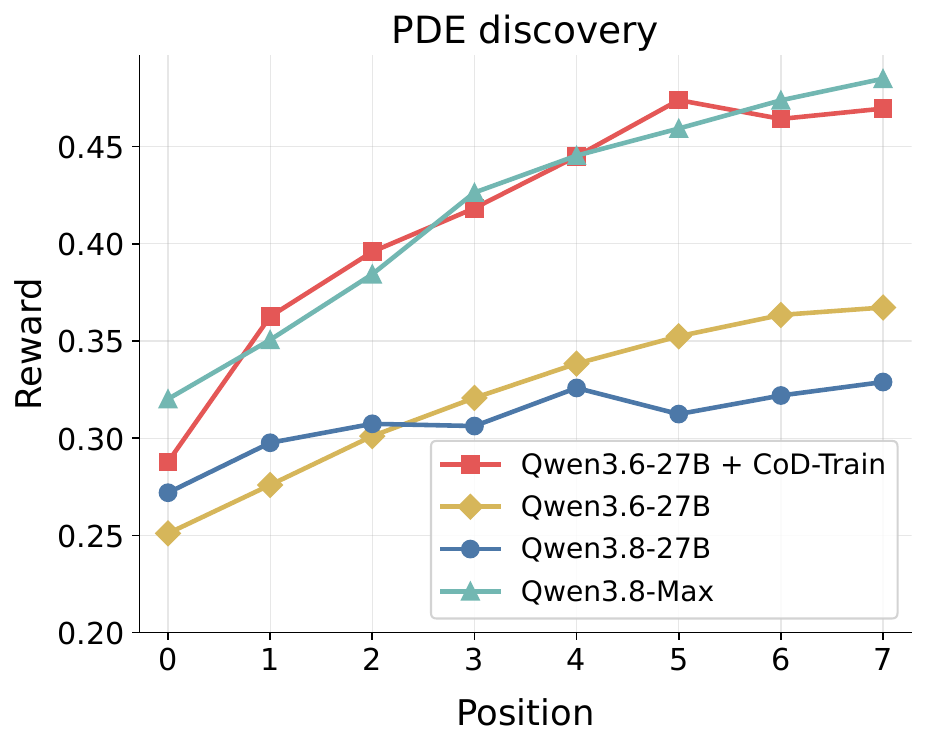}%
    \includegraphics[width=.31\textwidth]{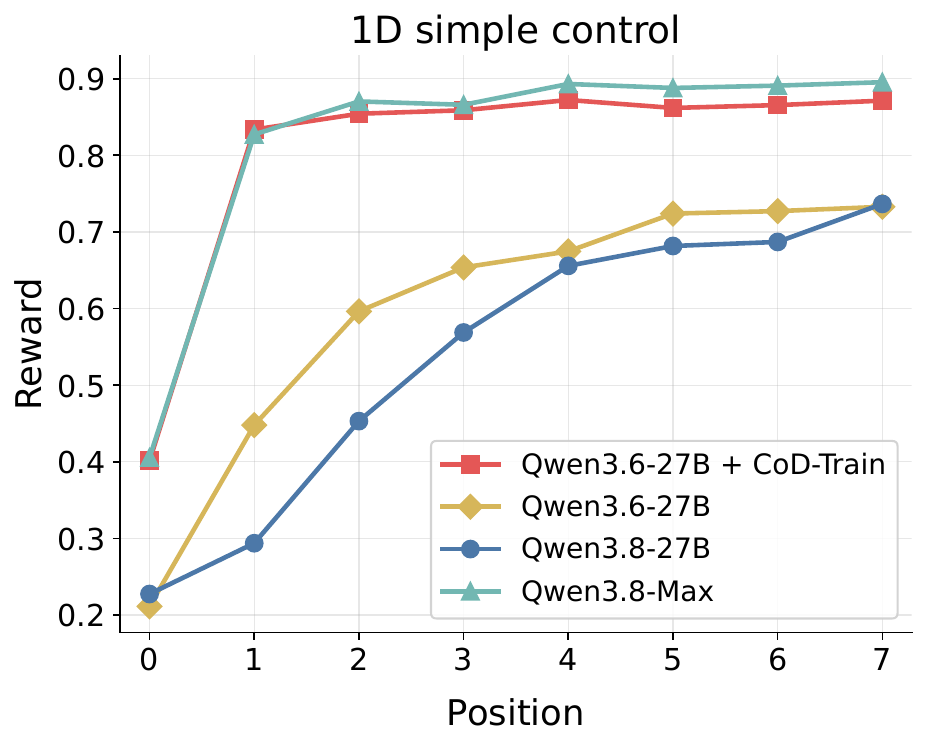}%
    \includegraphics[width=.31\textwidth]{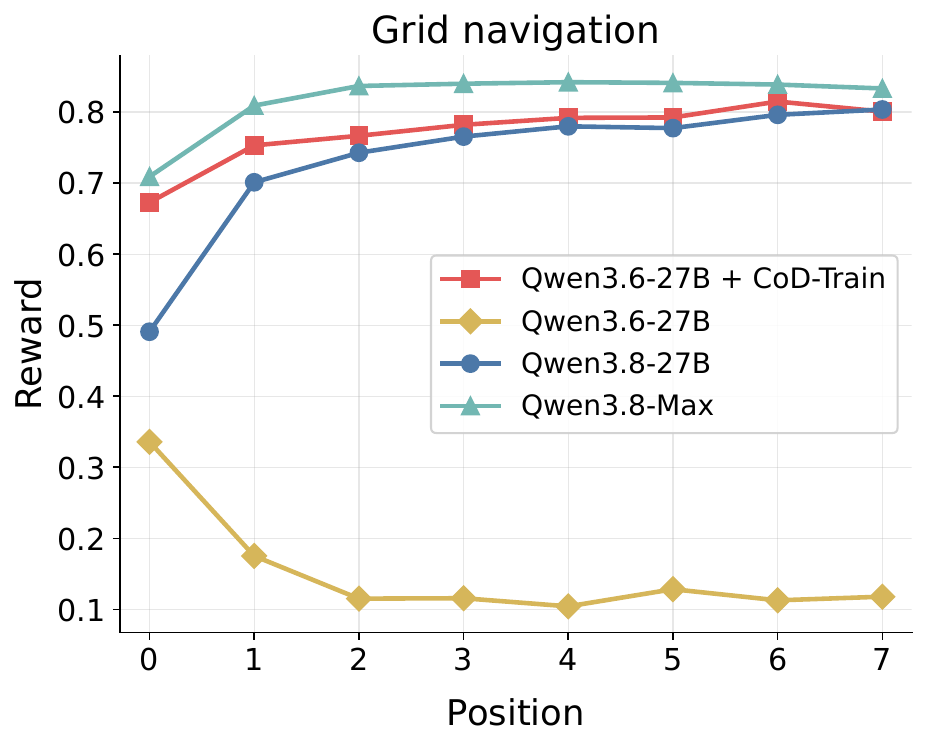}

    \vspace{-0.2em}
    \caption{Empirical comparison between a \cod-trained checkpoint (initialized at \qwensix) and frontier Qwen3.8 models.
    The X-axis denotes the position index of a task within a stream of tasks in the same environment;
    the \cod-trained 27B model learns to adapt efficiently at deployment and thereby achieves higher rewards in later tasks, 
    significantly outperforming \qwentpeseven and closely matching the performance of \qwentpemax.
    Further details about these results can be found in Section~\ref{sec:implementations_experiments}.
    }
    \label{fig:exp_27b_compare_qwen38}

\end{figure}

\vspace{0.5em}

\section{Introduction}
\label{sec:intro}

Large language models (LLMs) have progressed rapidly in the past few years,
evolving from chatbots to reasoning models,
and now serving as the backbones for numerous AI agents in real-world applications.
We envision that in the near future, further progress in LLMs will bring them to \emph{long-lifecycle agentic deployment} in diverse environments and \emph{continual learning at test time}.

To achieve this vision, an AI agent should be able to ``\underline{Co}nnect the \underline{D}ots'' (\cod).
More formally, we define the \cod \emph{meta-capability} as follows:
during long-lifecycle deployment in an \emph{environment},
an AI agent continuously solves a sequence of different-but-related \emph{tasks},
while proactively and deliberately exploring the environment and self-updating its \emph{context} about it, 
thereby facilitating more effective task solving within the same environment in the future.
In contrast to domain-specific capabilities (such as math problem solving and competitive coding),
\cod is a meta-capability that can potentially generalize across diverse environments from different domains,
enabling the AI agent to self-improve efficiently during deployment
and thereby enhancing its task-solving performance \citep{Silver2025,rafiee2026enactiveartificialintelligence}.
Frontier LLMs today still fall short in this regard:
with limited \cod meta-capability, an LLM-based agent may easily get lost in environments with underspecification \citep{laban2026llms,wei2025groundedrealitylearningdeploying}, and often requires human-crafted agent scaffolds with sophisticated design in order to operate stably over a long lifecycle.

We believe that LLMs need to be \emph{explicitly trained} for the \cod meta-capability, ideally through \emph{end-to-end reinforcement learning} (RL) in environments with clean reward signals.
Indeed, while early LLMs exhibited certain chain-of-thought reasoning capabilities \citep{wei2022chain,kojima2022large} and agentic intelligence \citep{yao2023react},
reinforcement learning --- with {properly designed algorithms and environments} --- has proved effective in driving fundamental and generalizable improvements in these aspects \citep{openai-o1,deepseek-r1,kimiteam2025kimik15scalingreinforcement,qwen35blog,kimiteam2026kimik25visualagentic,glm5team2026glm5vibecodingagentic,jagadeesh2026reinforcementlearningbroadlypersistently,shenfeld2025rlsrazoronlinereinforcement,chu2025sftmemorizesrlgeneralizes};
we anticipate that the same should hold for eliciting the \cod meta-capability in LLMs.
On the other hand, standard task-by-task RL essentially trains an LLM to solve each task individually \emph{from scratch},
which appears misaligned with the \cod meta-capability targeted for long-lifecycle deployment.
This motivates us to design and implement a dedicated RL process for \cod;
in this work, we present our initial progress.

\paragraph{Contributions and organization.}

\begin{figure}
\centering
\includegraphics[width=\textwidth]{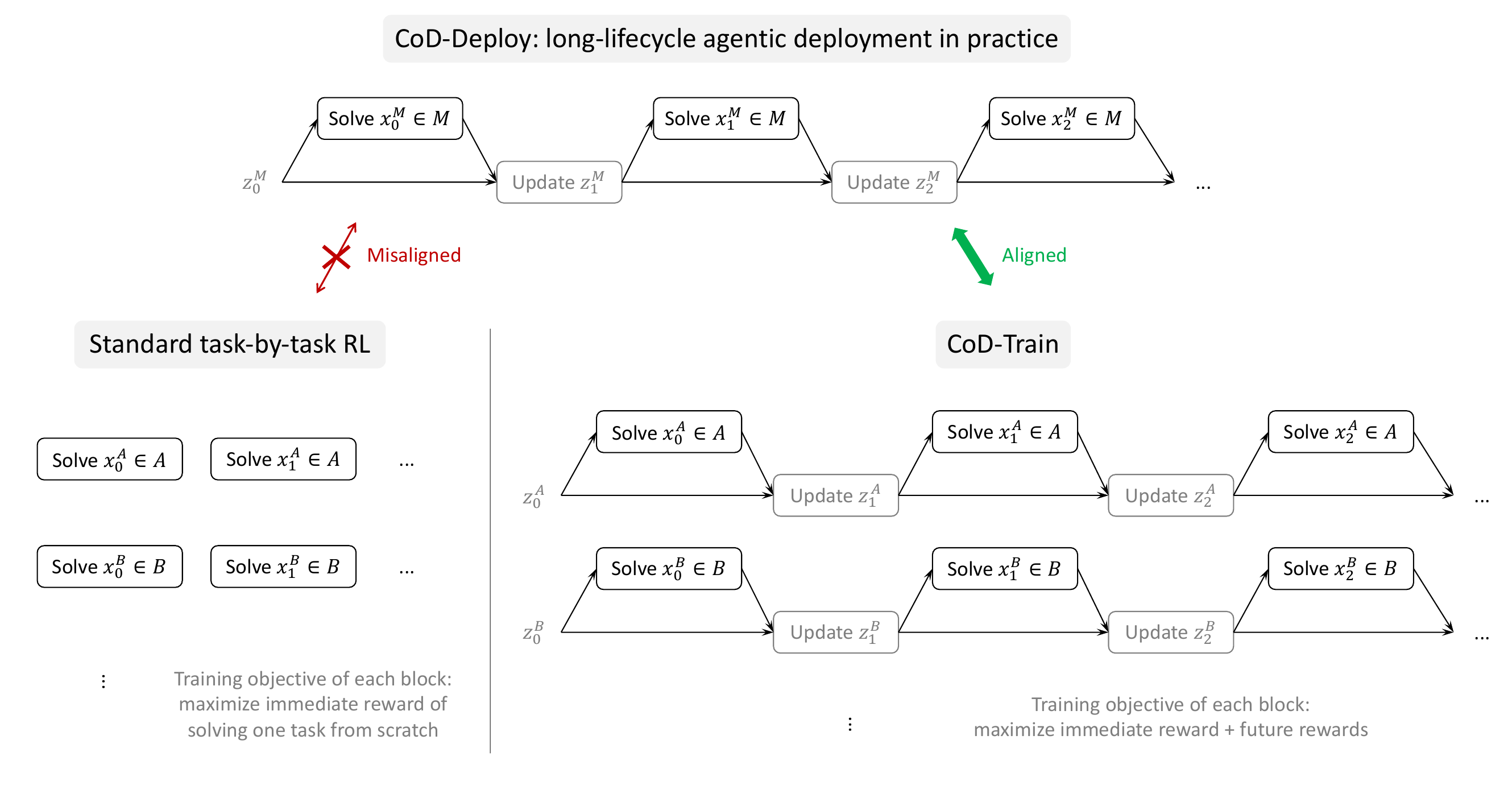}
\caption{A visualization of \coddeploy and \codtrain (compared with standard task-by-task RL),
where $A$ and $B$ denote different environments for training,
and $M$ denotes a new environment for deployment or evaluation.
Each block represents one rollout episode for solving a task $x_i$ (which could be in itself a long-horizon multi-turn task) or updating the agent's context $z_i$ about the current environment.
}
\label{fig:cod-viz}
\end{figure}

We begin by introducing a general \cod framework,
visualized in Figure~\ref{fig:cod-viz} and elaborated in Section~\ref{sec:framework}.
\coddeploy represents an abstraction of long-lifecycle deployment of an AI agent in an environment,
which interleaves \emph{episodes} for solving new tasks and for self-updating the agent's context about the environment.
\codtrain denotes the corresponding RL post-training process, whose rollout pattern exactly matches that of \coddeploy.

One major challenge lies in RL algorithm design, and in particular,
credit assignment across solve-task and update-context episodes in a long state-action sequence.
We adopt the classical dynamic-programming principle \citep{dynamicprogramming,sutton1998reinforcement}: the objective of each episode is to maximize not only the immediate reward, but also future rewards.
Another major challenge is the choice of appropriate tasks and environments \citep{huang2025environmentscalinginteractiveagentic}.
For training, we need environments that properly \emph{incentivize} self-updating an agent's context about the environment --- e.g., revealed information that was initially hidden, or task-solving skills with validated efficacy --- and solving new tasks conditioned on that context rather than from scratch.
For evaluation, we need environments that faithfully measure progress in improving the \cod meta-capability of LLMs.
The \cod framework therefore calls for dedicated task and environment design,
as standard RL datasets and LLM benchmarks are not necessarily suitable for these purposes.

In an effort to tackle these challenges, we present in Section~\ref{sec:implementations_experiments} our concrete implementations of the \cod framework,
which include RL algorithms and infrastructure, along with environments and tasks tailored to \cod.
Our empirical results validate the efficacy of end-to-end RL for eliciting the \cod meta-capability.
For example, when applying \codtrain to \qweneight on \flobs environments 
containing clues that are necessary for successful task solving but initially hidden,
the agent's success rate in solving a task \emph{from scratch} rises only from 18\% to 45\% (approximately) due to information-theoretic limits,
yet its success rate in solving the fourth task within a sequence --- \emph{conditioned on its self-updated context} --- increases substantially from 28\% to 76\%.
Moreover, the elicited \cod meta-capability demonstrates potential for in-domain and cross-domain generalization, 
as well as generalization to Ralph-loop settings \citep{ralphloop},
as validated by out-of-distribution evaluation.
Similar observations can be made as we scale up our experiments to larger LLMs, more challenging environments, and mixed-domain training by multi-teacher on-policy distillation \citep{lu2025onpolicydistillation,agarwal2024onpolicy};
notably, initialized at \qwensix, a \cod-trained model can significantly outperforms \qwentpeseven and closely matches the performance of \qwentpemax in \cod settings with simulated science and engineering environments.

During our development of \cod, we connect several lines of prior work,
including research and practice of ``lifelong agents'' \citep{lifelongagent}, meta reinforcement learning \citep{duan2016rl2fastreinforcementlearning}, inference scaling \citep{ralphloop}, among others.
A brief review of these lines, along with their connections to and differences from \cod, can be found in Section~\ref{sec:related_works}.

Finally, while this work provides proof-of-concept implementations of the \cod framework and presents promising empirical results,
many research questions remain open and call for further work,
e.g., refining the design of RL algorithms and environments dedicated to \cod,
continuing to scale up \cod along various dimensions,
and integrating \cod into existing LLM post-training pipelines;
detailed discussions can be found in Section~\ref{sec:discussions}.

\section{The CoD framework}
\label{sec:framework}

In this section, we introduce the \cod framework --- visualized in Figure~\ref{fig:cod-viz} ---
that aims to elicit the aforementioned \cod meta-capability in LLMs via RL post-training.
\begin{itemize}
\item \coddeploy is an abstraction of long-lifecycle agentic deployment in an environment $M$, 
which interleaves episodes for solving a task $x_i^M$ and for updating the agent's context from $z_{i}^M$ to $z_{i+1}^M$.
This process can be understood as (gradient-free) online learning via trial-and-errors.
\item \codtrain represents the rollout mechanism in RL post-training dedicated to \cod, and the corresponding RL algorithm for training model weights.
Note that the rollout pattern in \codtrain is simply the same as in \coddeploy, covering diverse environments $A, B, \dots$ for generalization.
\end{itemize}
From the perspective of \emph{hierarchical reinforcement learning},
\codtrain goes up by one level in the hierarchy, 
compared to standard task-by-task RL.
Indeed, the definition of one rollout trajectory in LLM-RL has evolved from a \emph{sequence of tokens} for a single-turn task (e.g., RLHF \citep{Ouyang2022,bai2022training} and reasoning-oriented RLVR \citep{openai-o1,deepseek-r1}), to a \emph{sequence of turns} for a long-horizon agentic task \citep{zhou2024archer,glm5team2026glm5vibecodingagentic,kimiteam2026kimik3openfrontier};
\codtrain takes it one step further, to a \emph{sequence of tasks} for long-lifecycle agentic deployment in an environment.

\paragraph{End-to-end reinforcement learning.}

Given the problem formulation in \codtrain, we aim for effective end-to-end RL training. 
One crucial aspect here is \emph{credit assignment} for a long state-action sequence consisting of multiple solve-task and update-context episodes,
each of which might already contain multiple turns of agent-environment interaction and thus a large number of tokens.
Fortunately, we have one critical observation: suppose that there is (at least) an outcome reward for solving each task, then the total number of reward values in the end-to-end state-action sequence will grow proportionally with the number of tasks, i.e., the \emph{density} of reward signals is independent of the total number of tasks.

We follow the fundamental dynamic-programming principle \citep{dynamicprogramming,sutton1998reinforcement} for credit assignment in \codtrain: each episode should aim for maximizing not only the immediate reward, but also future rewards.
In an update-context episode, the agent performs deliberate condensation of its previous experiences and incorporates it into the updated context, which should contribute to better performance in solving future tasks within the same environment.
In a solve-task episode, the agent takes action --- conditioned on its self-updated context --- while balancing exploitation and exploration,
i.e., trying to solve the current task while simultaneously exploring the environment and collecting new information / discovering task-solving skills that are worthy of being incorporated into the updated context.

\paragraph{Tasks and environments for training and evaluation.}

Choosing the right tasks and environments for \cod decides whether \codtrain can effectively elicit the targeted meta-capability as we desire,
and whether evaluation results in \coddeploy align well with the LLM's performance in real-world long-lifecycle scenarios and faithfully reflect the progress of LLM development.
\begin{itemize}
\item Ideal tasks and environments for training should ensure that an agent has sufficient \emph{incentives} for iteratively updating its context about the environment and transferring it across different tasks. 
This is true, for example, if successful task solving requires certain information about the environment that the agent cannot possibly access prior to deployment;
in this case, solving each task independently would hit an information-theoretic performance limit, and the agent is thus incentivized to continuously accumulate knowledge about the environment and transfer it across tasks.
In other scenarios, the agent might be incentivized to propose and try out different task-solving skills, observe environmental feedback, and then focus on skills with validated efficacy for future tasks within the same environment.
The distribution of training tasks plays a crucial role in incentivizing the trained model to make context-aware decisions and mitigating incorporation of undesirable priors or bias into the trained model; a minimal demonstration and theoretical understanding of this aspect can be found in Appendix~\ref{sec:appendix_dynamics}.
\item Similarly, ideal tasks and environments for evaluation need to align well with long-lifecycle agentic deployment in challenging real-world scenarios, 
where the best overall performance can be achieved only if the agent proactively maintains its context about the environment and solves new tasks conditioned on it (rather than from scratch).
\end{itemize}
Nonetheless, more generic tasks and environments might still be worthy of consideration:
\begin{itemize}
\item For training, if it is possible in theory to achieve maximum reward for each task by solving it from scratch (without any extra context), then there is a risk that \codtrain might degenerate into standard task-by-task RL.
Yet another possibility is that \codtrain might exhibit different learning dynamics than those of standard RL (e.g., due to the natural integration of text-based environment feedback into the rollout process), which could still lead to acquisition of the \cod meta-capability, better out-of-distribution generalization, and/or higher sample efficiency than standard RL.
\item For evaluation, any existing benchmark can be converted to a \cod counterpart by regarding the whole benchmark (or a certain subset of it) as a collection of tasks from the same environment.
Applying \coddeploy in this setting could lead to better context --- generated automatically by the agent itself with a handful of trial-and-errors --- that improves the agent's performance on the same benchmark.
The \cod meta-capability also plays a role in a Ralph-loop setting \citep{ralphloop} where an agent attempts repeatedly to solve the same task, or within a single episode for solving a long-horizon task that requires delicate context management.
\end{itemize}

\paragraph{Primary research questions.}

As we set out on implementing the \cod framework, several research questions of interest naturally arise, for example:
\begin{itemize}
\item What is the suitable RL algorithm for \codtrain, e.g., how to calculate advantages and gradient updates? Is it possible to achieve more effective and efficient training by taking advantage of the particular rollout pattern in \codtrain? Conversely, is it possible to design a better rollout pattern that allows more effective RL training?
\item What are some concrete examples of appropriate tasks and environments for training and evaluating the \cod meta-capability in LLMs? To what extent will the learned \cod meta-capability generalize? Curiously, if \codtrain is applied to game or synthetic environments (which essentially allow infinite training data), will the elicited \cod meta-capability generalize to real-world scenarios, e.g.,
a personal assistant that can continuously accumulate knowledge about a specific user and thereby improve its quality of service over time,
or a coding agent that can incrementally develop a repository and maintain its health over an extended period (like an expert engineer does)?
\item What happens when \codtrain (resp.~\coddeploy) is applied with more generic tasks and environments for training (resp.~evaluating) an LLM? What benefits would arise from the natural integration of text-based environment feedback in \codtrain and \coddeploy? Could \codtrain automatically decide what should be learned into model weights and what into context, leading to better learning efficiency and/or generalization?
\end{itemize}
We provide our preliminary answers to some of these questions in the next section.

\section{Proof-of-concept implementations and experiments}
\label{sec:implementations_experiments}

In this section, we present our implementations of the \cod framework,
as well as preliminary empirical results that validate the efficacy of end-to-end RL in eliciting the \cod meta-capability.

\subsection{RL algorithms and infrastructure}
\label{subsec:rl_algorithm_infrastructure}

Our RL algorithms and infrastructure tailored to \cod are elaborated below.

\paragraph{Reward design.}

We consider tasks with outcome rewards for simplicity in our current implementations.
We thus assign a task-wise outcome reward to each solve-task episode within a rollout trajectory in \codtrain.
For each update-context episode, we assign only a small format reward, since there is no other direct reward signal for it.

\paragraph{RL algorithm design.}

\begin{figure}
    \centering
    \includegraphics[width=\textwidth]{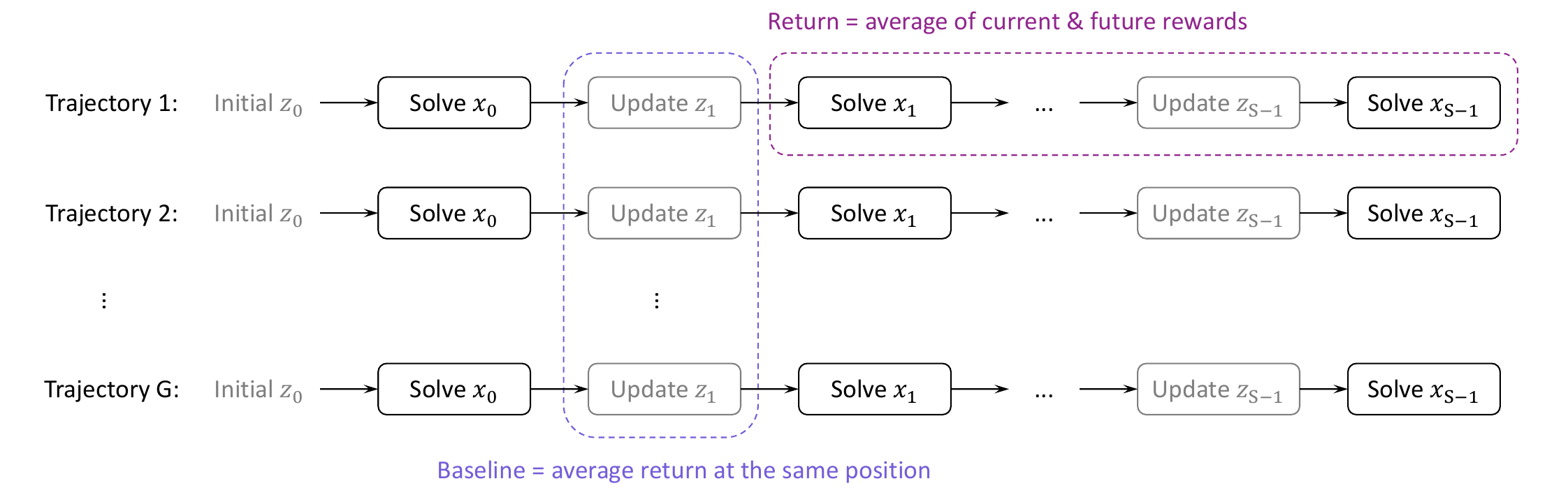}
    \caption{Advantage calculation in our RL algorithm tailored to \codtrain.}
    \label{fig:viz_cod_advantage}
\end{figure}

We adopt a GRPO-style \citep{deepseek-math} training algorithm in \codtrain,
which requires no critic model and assumes access to multiple rollout trajectories for the same sequence of tasks.
Note that GRPO originally assumes a single outcome reward for each rollout trajectory,
whereas in \codtrain, each rollout trajectory contains a sequence of reward values for multiple episodes.
To account for this, we define the return of each episode as the mean reward of the current episode plus future solve-task episodes.
Then, we regard episodes at the same position in its corresponding rollout trajectory as one group, using their averaged return as the baseline for advantage calculation.
Figure~\ref{fig:viz_cod_advantage} presents a visualization of credit assignment and advantage calculation in our RL algorithm tailored to \codtrain.
To resolve training instability observed in our early experiments, we further add a heuristic augmentation that balances positive and negative gradients by re-weighting training samples adaptively.
See Appendix~\ref{subsec:complete_rl_algorithm} for full details of our final RL algorithm.

\paragraph{Multi-teacher on-policy distillation (OPD).}

One straightforward approach for integrating environments from diverse domains into \codtrain is to run a single RL process with mixed environments.
Another approach --- also supported by our implementations --- is to first train an expert model for each domain by RL, and then merge their abilities into a single model by multi-teacher OPD \citep{lu2025onpolicydistillation,agarwal2024onpolicy}.
As will be demonstrated in Section~\ref{subsec:exp_qwen36_27b}, the OPD approach performs well in terms of sample efficiency, flexibility, and training stability in our \cod settings.

\paragraph{RL infrastructure.}

We build on \trinity \citep{trinity},
an LLM-RL framework with a modular and decoupled design that facilitates convenient customization of various components involved in the RL pipeline.
We implement an environment-wise meta-workflow in the form of \coddeploy, so that any task and its corresponding task-wise rollout workflow can be readily plugged in after minor modifications.

\paragraph{A minimal update-context workflow.}

While our infrastructure supports generic update-context workflows, we focus on a minimalistic instantiation of cross-episode context --- also termed ``\emph{hint}'', which is essentially a single piece of text that will be appended to the system prompt in a solve-task episode --- in our major experiments.
We find it sufficient for proof-of-concept implementations and experiments of the \cod framework;
one important direction for future work is to generalize the update-context mechanism, e.g., involving persistent memory \citep{dorovatas2026positionmodularmemorykey,suzgun2025dynamiccheatsheettesttimelearning,zhou2025mem1learningsynergizememory,zhang2026memoryactionautonomouscontext,yan2026memoryr1enhancinglargelanguage} or Agent Skills \citep{anthropicskills,zhang2026skillflowbenchmarkinglifelongskilldiscovery,li2026skillsbenchbenchmarkingagentskills,yao2026skillriseagenticreinforcementlearning}, which could be crucial for performance improvements in more challenging scenarios.

\subsection{Tasks and environments}

As explained in Section~\ref{sec:framework}, generic tasks and environments for standard task-by-task RL are not necessarily appropriate for training and evaluation of the \cod meta-capability.
We thus design and implement dedicated environments for our investigation of \cod, which are elaborated in the following.

\subsubsection{Toy environments for \qweneight experiments}

\flobs is analogous to the classic \frozenlake game \citep{brockman2016openaigym,towers2026gymnasium}, but with obscurity in the environments.
In a standard \frozenlake task, an agent navigates on a 2D grid map by choosing one action from moving up/down/left/right at each step, and needs to reach the goal position within a limited number of steps while avoiding holes in the map.
The key difference in \flobs is that the action space now becomes A/B/C/D, and for each new environment, the {mapping} between A/B/C/D and up/down/left/right is randomly permuted and unknown \textit{a priori}. 
This imposes an information-theoretic limit on the success rate that an agent can possibly achieve when solving each task individually without additional context\footnote{For example, if the starting position of the agent is surrounded by three holes, then without prior information about the ground-truth action mapping, the agent will fail by falling into a hole in the first step with probability $75\%$ on average.
Similarly, if the agent's action budget is only slightly above the length of the shortest path between the starting and targeted positions, then the agent might not have the chance to figure out the action mapping through trial-and-errors within the same episode.
}.
The agent, when deployed in a new environment, is thus \emph{incentivized} to maintain its context and transfer it across episodes when solving a sequence of tasks within this environment.
A good ``hint'' for a \flobs environment should thus contain the ground-truth mapping between A/B/C/D and up/down/left/right, derived by the agent itself from previous solve-task episodes.
Despite its simplicity, \flobs contains the fundamental elements of long-lifecycle agentic deployment, making it a minimal sanity-check testbed for \cod,
e.g., for validating an RL algorithm dedicated to \codtrain, or evaluating generalization of an LLM's \cod meta-capability.

\alchrand is another game scenario, designed by the same principles and inspired by the classic \alchemy benchmark \citep{wang2021alchemy}.
Each \alchrand environment has its own set of elements and recipes (i.e., which pair of elements can be combined to synthesize another element);
these are randomly initialized and then fixed for each environment.
In each task, an agent is asked to synthesize a targeted element within a limited number of steps,
without any prior knowledge about the ground-truth recipes initially when deployed in a new environment.
A good hint in the \alchrand domain should ideally contain valid recipes and failed combinations --- discovered by the agent itself in previous solve-task episodes --- which facilitate more efficient searching when solving a new task; it might also contain task-solving skills tailored to this domain, e.g., advanced strategies for systematically deciding which combination to try out at each step.
Compared with \flobs, \alchrand allows more diverse environments, a wider range of configurable difficulty for the tasks, and richer behaviors by the agent.

\terminal is a simulation scenario that tests an agent's capabilities in daily tasks relevant to Linux / MacOS / Windows terminals, such as file manipulation and transferring.
The agent is asked to solve each task within a limited number of steps; for each step, the agent observes the current environment state, executes a terminal command, and receives feedback from the environment.
Since the optimal approach for solving similar tasks can vary depending on characteristics of the environments (e.g., operating systems, or sizes of the files to be manipulated), it is generally useful for the agent to explore the current environment and unveil its characteristics while solving a specific task.

\subsubsection{Science and engineering environments for \qwensix experiments}

\pde is about solving PDE inverse problems from sampled data.
When deployed in a \pde environment, the agent's goal is to infer the underlying PDE formula of a system.
In each solve-task episode, the agent is allowed to access a new time-series trajectory evolving by the ground-truth PDE formula (starting from a random initial condition), sample datapoints on a limited number of timestamps and locations chosen by itself, call a sparse-regression tool \citep{brunton2016discovering,rudy2017data} acting on the accumulated data, analyze the results, and propose its updated PDE formula.  
The reward for each solve-task episode is determined by the distance between the agent's answer and the ground-truth PDE formula, which is only used in RL training and never revealed in the rollout trajectory (since in \coddeploy it is impossible to calculate this reward when the ground-truth formula is unknown).

\control is about designing good policies for controlling 1D linear dynamical systems.
When deployed in a \control environment --- corresponding to one system with noisy dynamics and initially unknown parameters --- the agent's goal for each task is to design a control policy that drives the system from an initial state to a targeted state.
In each solve-task episode, the agent proposes a control policy and then observes the resulting time-series trajectory, which allows it to infer the underlying parameters of the dynamical system and thus improve its control policies in later tasks.
The reward for each solve-task episode is determined by the error between the terminal and targeted states, plus accumulated control efforts.

\navigation is about finding short paths in 2D grids.
When deployed in a \navigation environment --- corresponding to a 2D grid with fixed but initially unknown cost values, one for each cell in the grid --- the agent's goal for each task is to navigate from a start position to a targeted position within a limited number of vertical or horizontal moves.
The reward for each solve-task episode is determined by the cost values of the cells on the chosen path.
The cost values of cells near the chosen path are revealed by the end of one solve-task episode; 
the agent is thus incentivized to balance exploration and exploitation in its path-finding strategy, and utilize such revealed information for improving its performance in later tasks within the same environment.

\subsection{Empirical validation with \qweneight}

We present our preliminary empirical results that validate the efficacy of end-to-end RL for \codtrain, 
as well as its potential of out-of-distribution (OOD) generalization.
Two concrete settings of training are considered for our main experiments:
\begin{itemize}
    \item \textbf{Setting A}: \codtrain on the \flobs domain, with task sequences of length 4;
    \item \textbf{Setting B}: \codtrain on the mixture of \flobs and \alchrand domains, by a single RL process and also with task sequences of length 4.
\end{itemize}
For both settings, we start with the \qweneight model \citep{yang2025qwen3technicalreport}.
Other hyperparameters and configurations for the \codtrain experiments can be found in Appendix~\ref{subsec:exp_config}.
Prompt templates for solve-task and update-context episodes can be found in Appendix~\ref{subsec:prompt_templates}.
To validate in-domain OOD generalization, we evaluate the trained model checkpoints on the same training domains, but with harder environment instances (e.g., larger maps in \flobs, or a larger number of elements in \alchrand) and longer task sequences of length 8.
To validate cross-domain OOD generalization, we evaluate the trained model checkpoints on domains that were unseen during training.
We also investigate OOD generalization to Ralph-loop settings \citep{ralphloop}, where an agent repeatedly solves the same task in order to arrive at a better solution;
this can be regarded as a special case of \coddeploy, where a sequence of tasks in one specific environment happen to be exactly the same task, but solved by the agent with different context.

\begin{figure}
    \centering
    \includegraphics[width=.45\textwidth]{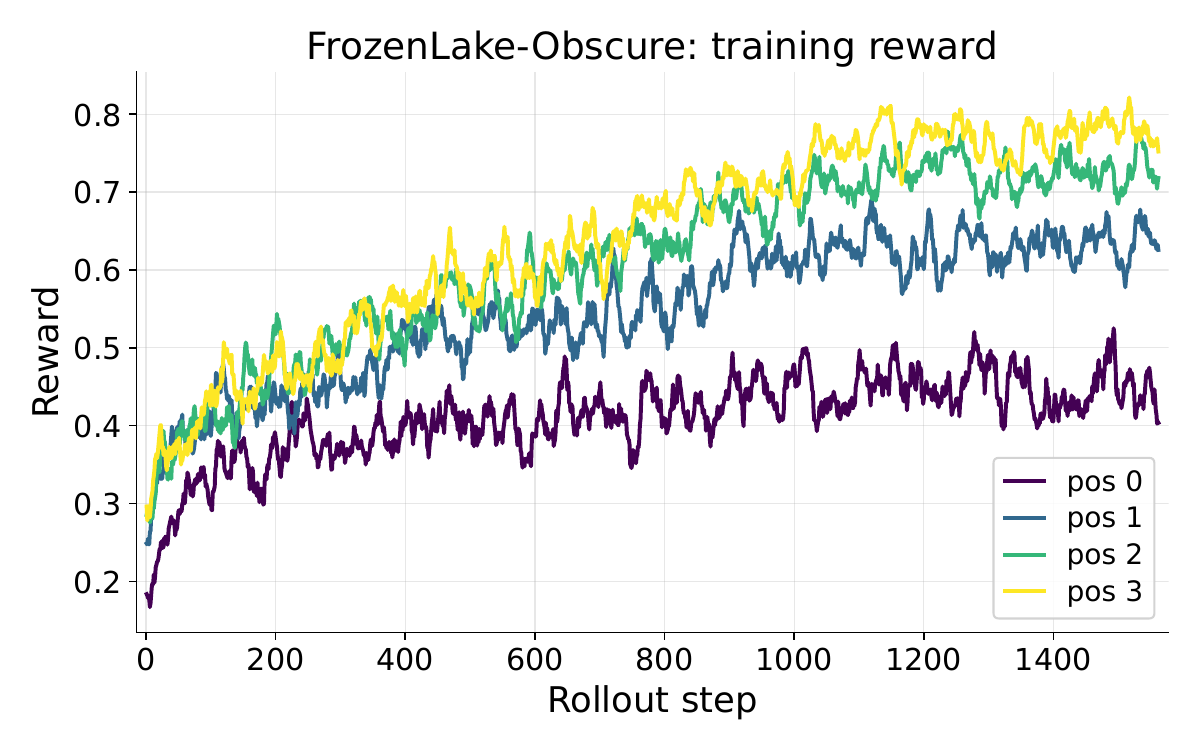} \\
    \includegraphics[width=.45\textwidth]{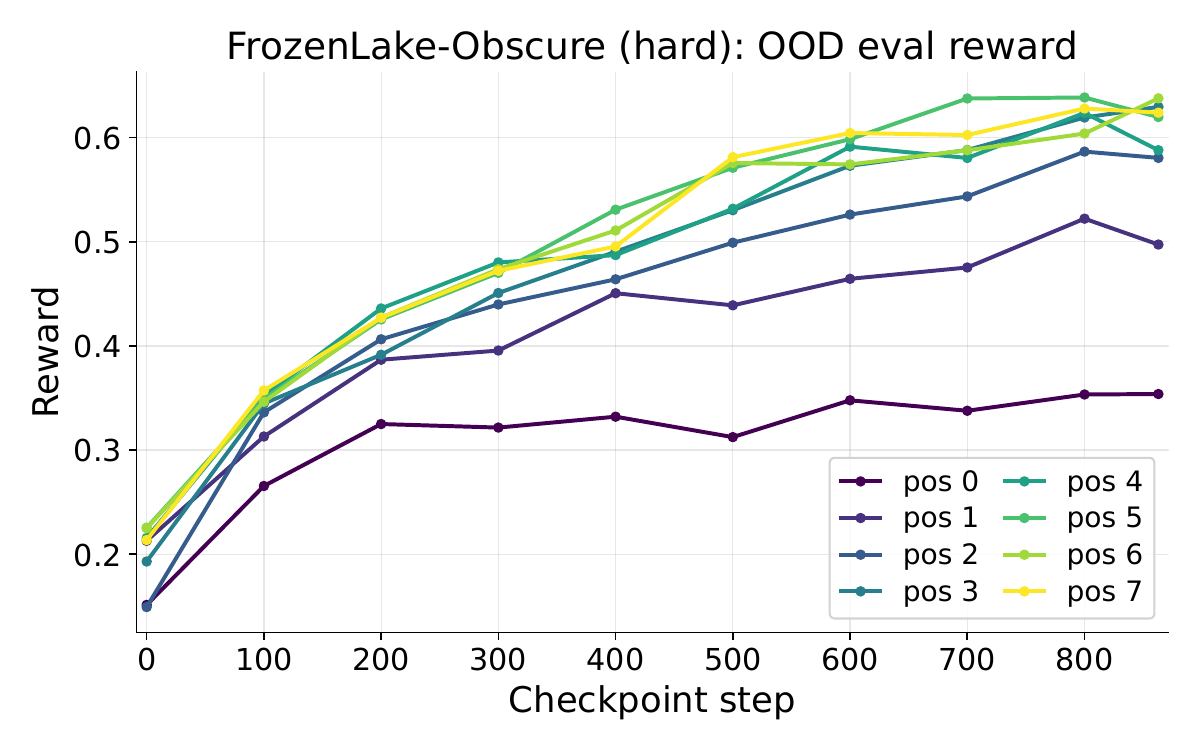}
    \includegraphics[width=.45\textwidth]{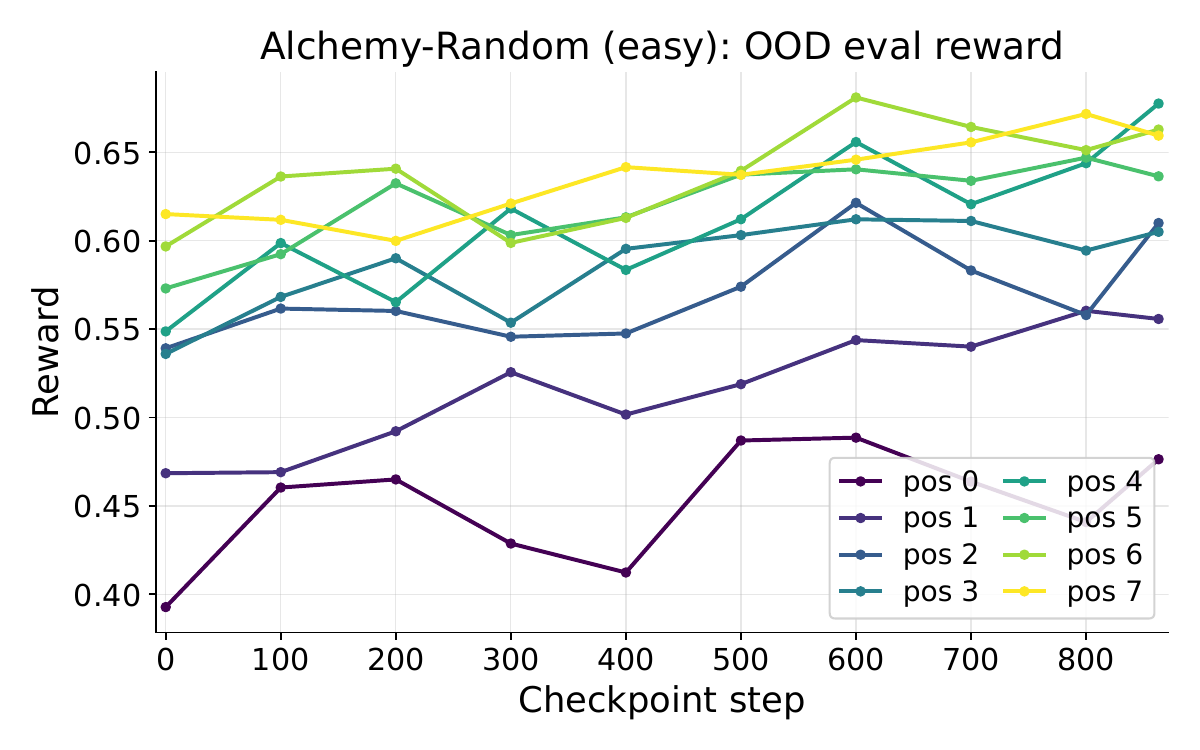} \\
    \includegraphics[width=.44\textwidth]{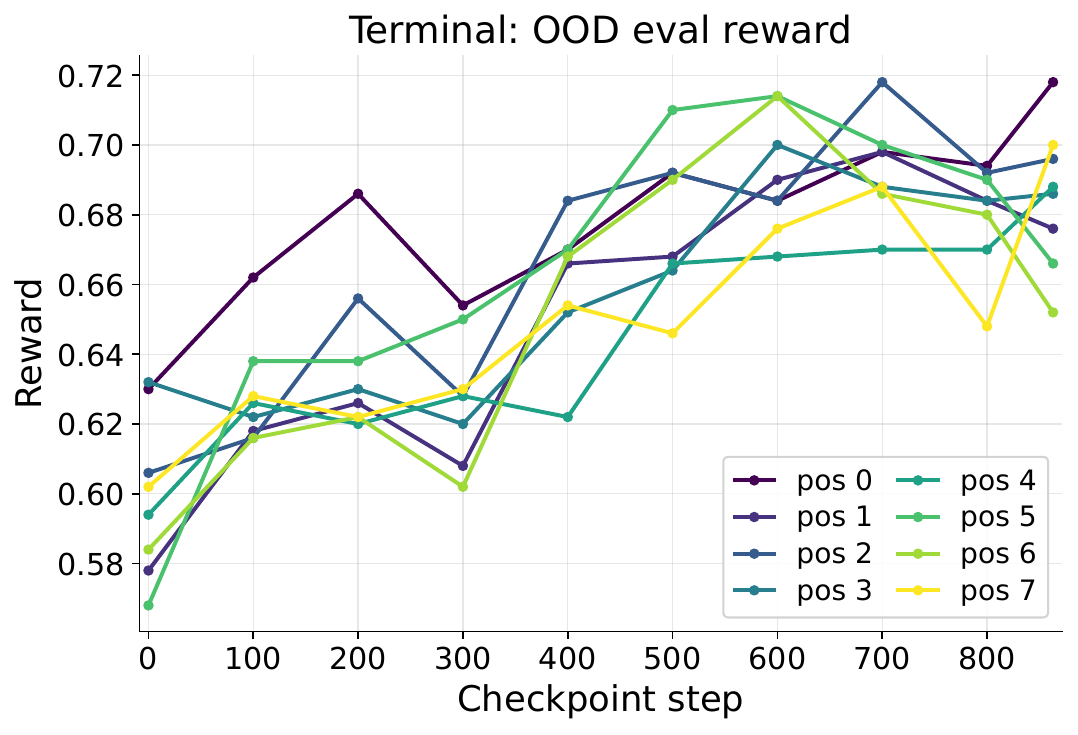}
    \includegraphics[width=.44\textwidth]{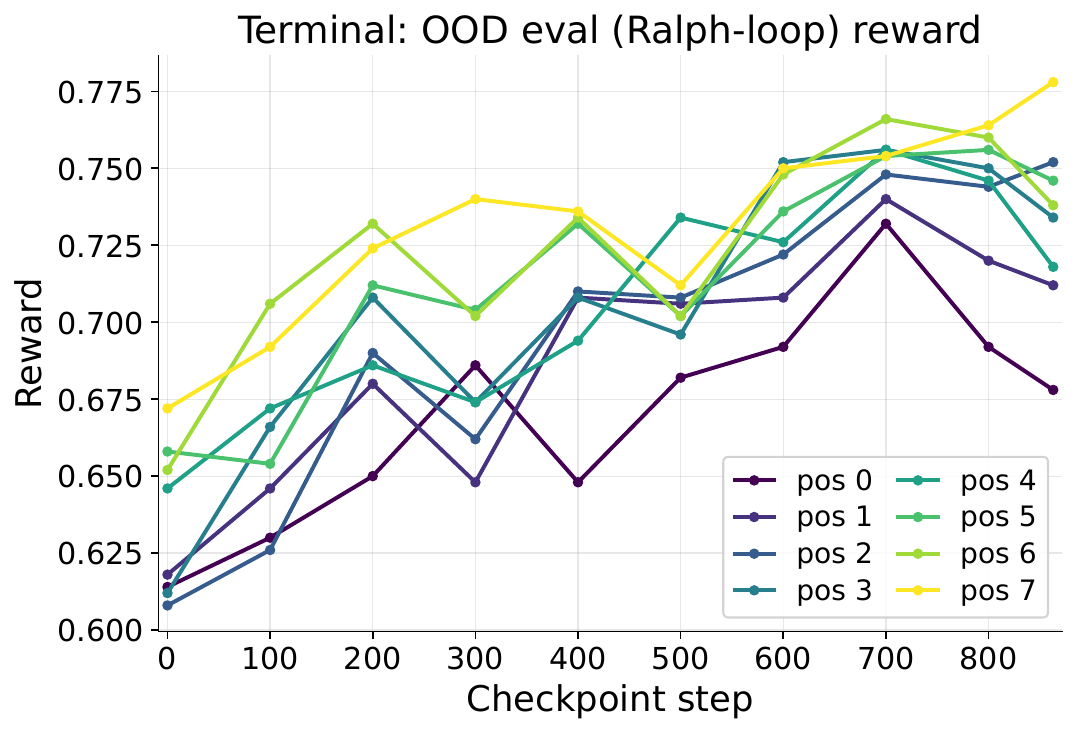}
    \caption{Empirical results for \qweneight in setting A, that is, \codtrain by RL on the \flobs domain.
    Training reward curves are smoothed with a running-average window of size 17.
    For evaluation curves, checkpoint step 0 corresponds to the initial \qweneight model.
    }
    \label{fig:exp_setting_A_results}
\end{figure}

\begin{figure}
    \centering
    \includegraphics[width=.45\textwidth]{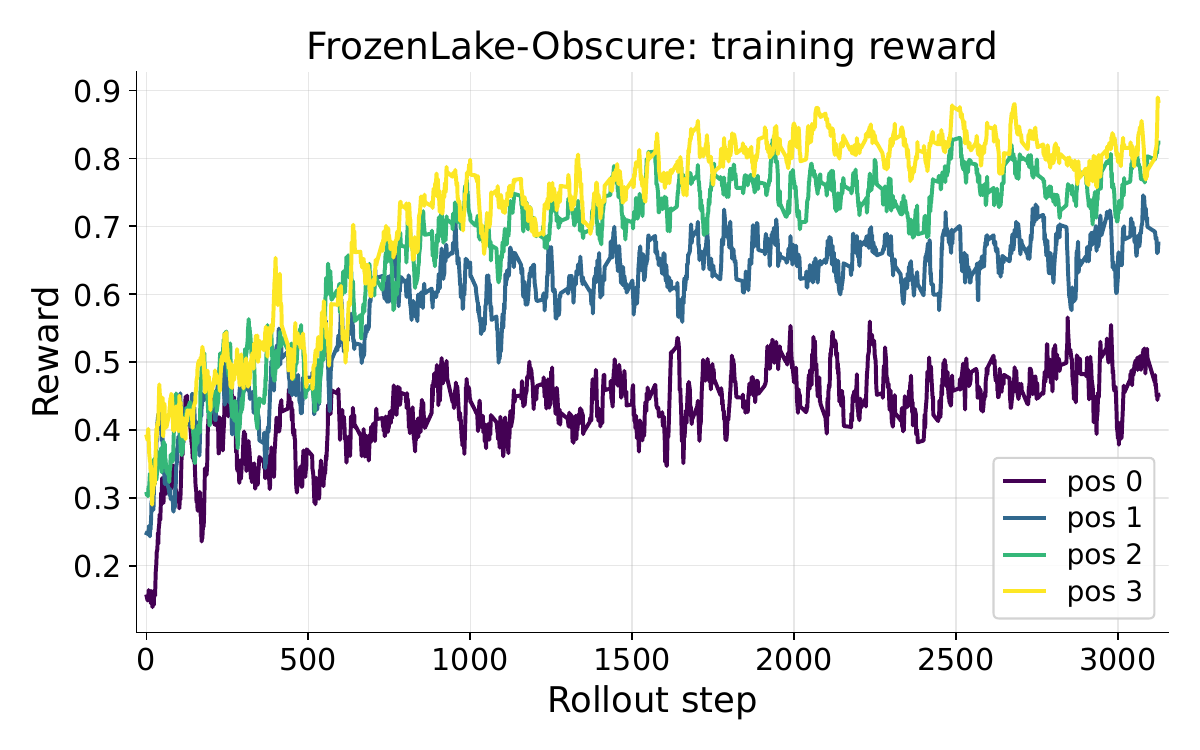}
    \includegraphics[width=.45\textwidth]{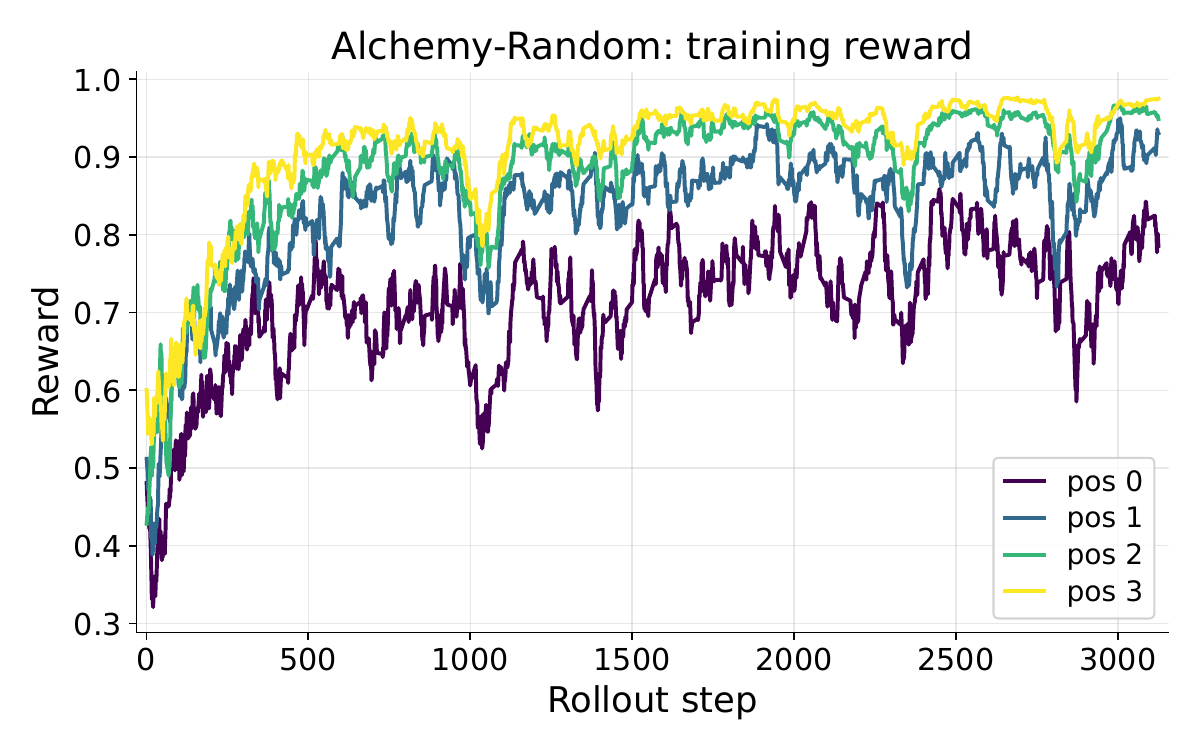} \\
    \includegraphics[width=.45\textwidth]{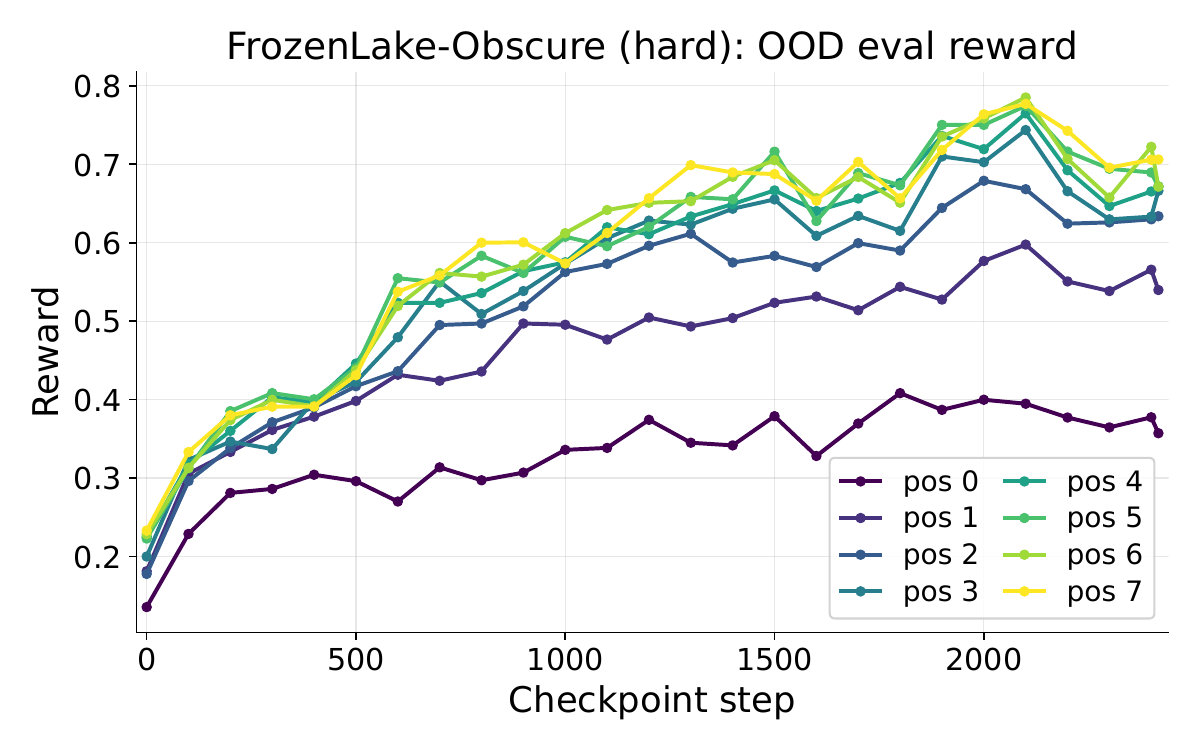}
    \includegraphics[width=.45\textwidth]{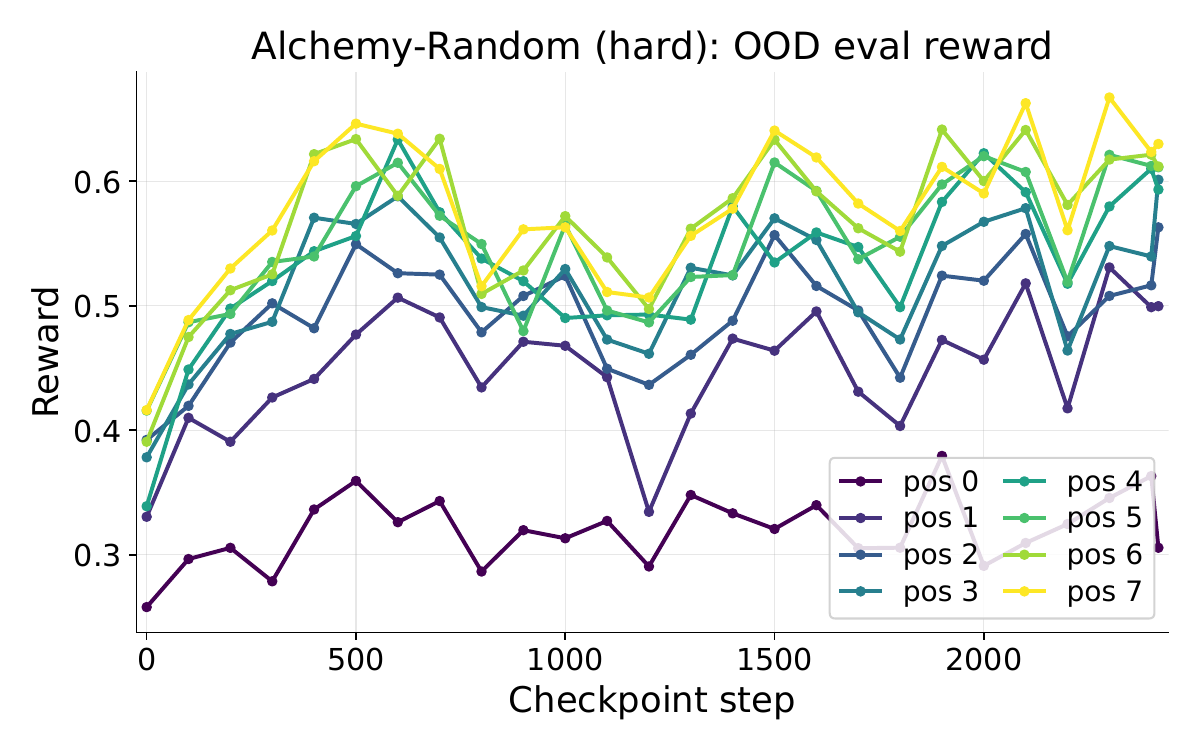} \\
    \includegraphics[width=.44\textwidth]{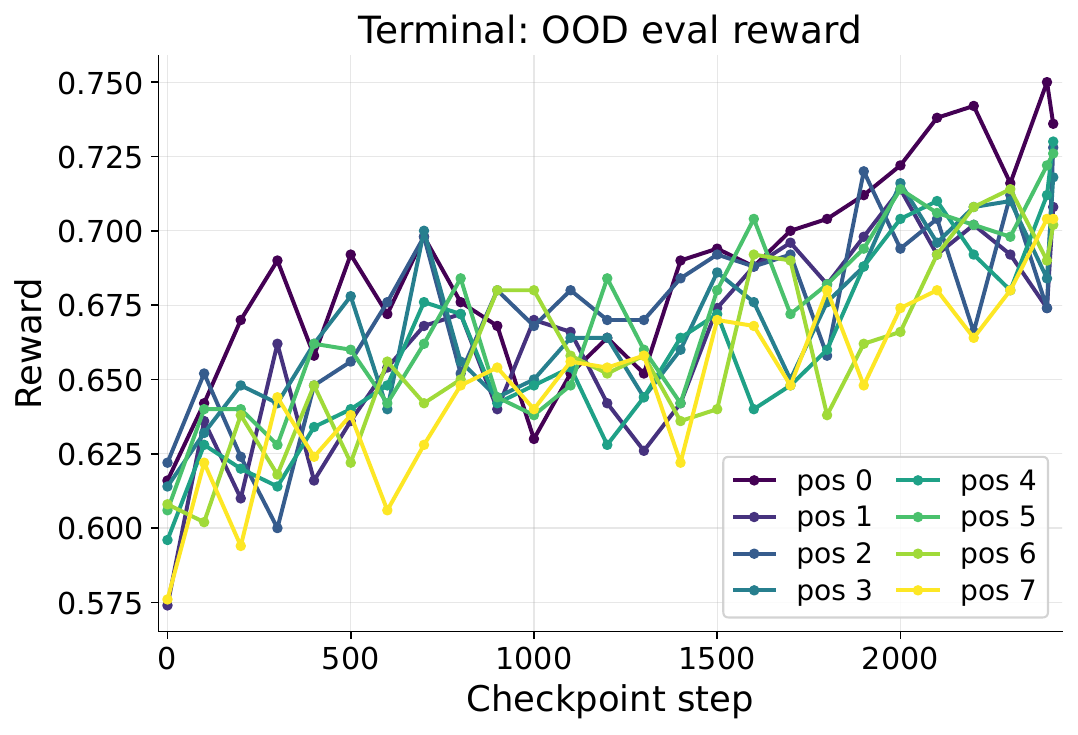}
    \includegraphics[width=.44\textwidth]{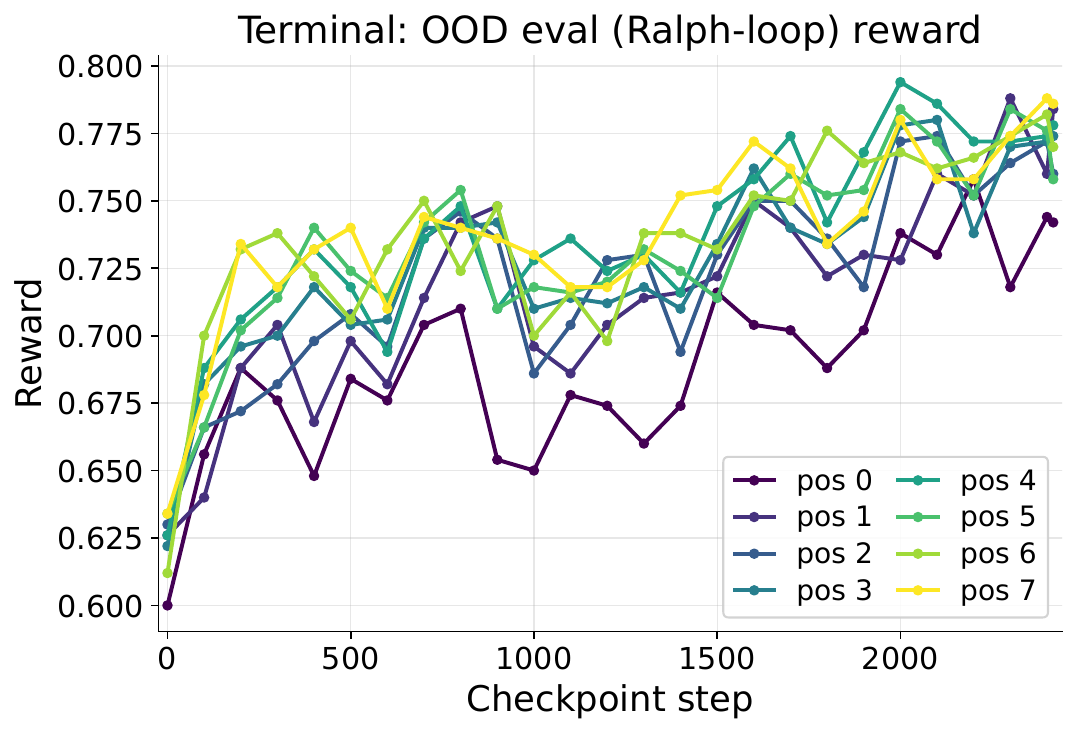}
    \caption{Empirical results for \qweneight in setting B, that is, \codtrain by RL on the mixture of \flobs and \alchrand domains.
    Training reward curves are smoothed with a running-average window of size 17.
    For evaluation curves, checkpoint step 0 corresponds to the initial \qweneight model.
    }
    \label{fig:exp_setting_B_results}
\end{figure}

\paragraph{Figure~\ref{fig:exp_setting_A_results}: results for setting A.}
We first observe that training reward curves grow with RL steps,
and moreover,
higher rewards can be achieved at later positions within a task sequence.
For example, the mean reward at position 0 (i.e., solving the initial task without any context about the current \flobs environment) grows from 0.18 to 0.45 during \codtrain,
whereas the mean reward at position 3 grows substantially from 0.28 to 0.76.
Reward curves for in-domain evaluation --- with harder environment instances and longer task sequences --- show similar results,
confirming the in-domain OOD generalization of \codtrain.
Finally, performance improvements can be observed for evaluation on \alchrand and \terminal (in both \coddeploy and Ralph-loop settings), which demonstrates a certain degree of cross-domain generalization\footnote{
Caution should be taken when interpreting the evaluation results on \terminal.
Notice that in the \coddeploy setting (with a sequence of different tasks), later tasks in the sequence show no performance gain over early ones, which is likely because in our current implementations, there is no close relation between different \terminal tasks within one sequence.
In comparison, for the Ralph-loop setting (with a sequence of repeated attempts for the same task), later episodes do achieve higher rewards than early ones.
We conjecture that the meta-capability of updating and utilizing the agent's context about the environment --- learned via \codtrain --- is also beneficial \emph{within} each solve-task episode, which might explain the upward reward curves for \terminal.
With that said, further investigation is needed for pinpointing the root cause of cross-domain performance improvements on \terminal after \codtrain with \flobs environments.
}.

\paragraph{Figure~\ref{fig:exp_setting_B_results}: results for setting B.}
One noticeable difference from the results of setting A is that training reward curves are less stable in setting B, reflected by minor performance degradation within the reward curves for \alchrand.
Similarly, evaluation reward curves for \alchrand grow rapidly in early steps, and fluctuate later on.
Interested readers might refer to Appendix~\ref{subsec:complete_rl_algorithm} for further discussions on the training stability of different RL algorithms.
Other than that, the results and conclusions from experiments in setting B are similar to those in setting A.

\subsection{Empirical validation with \qwensix}
\label{subsec:exp_qwen36_27b}

\begin{figure}
    \centering

    \includegraphics[width=.45\textwidth]{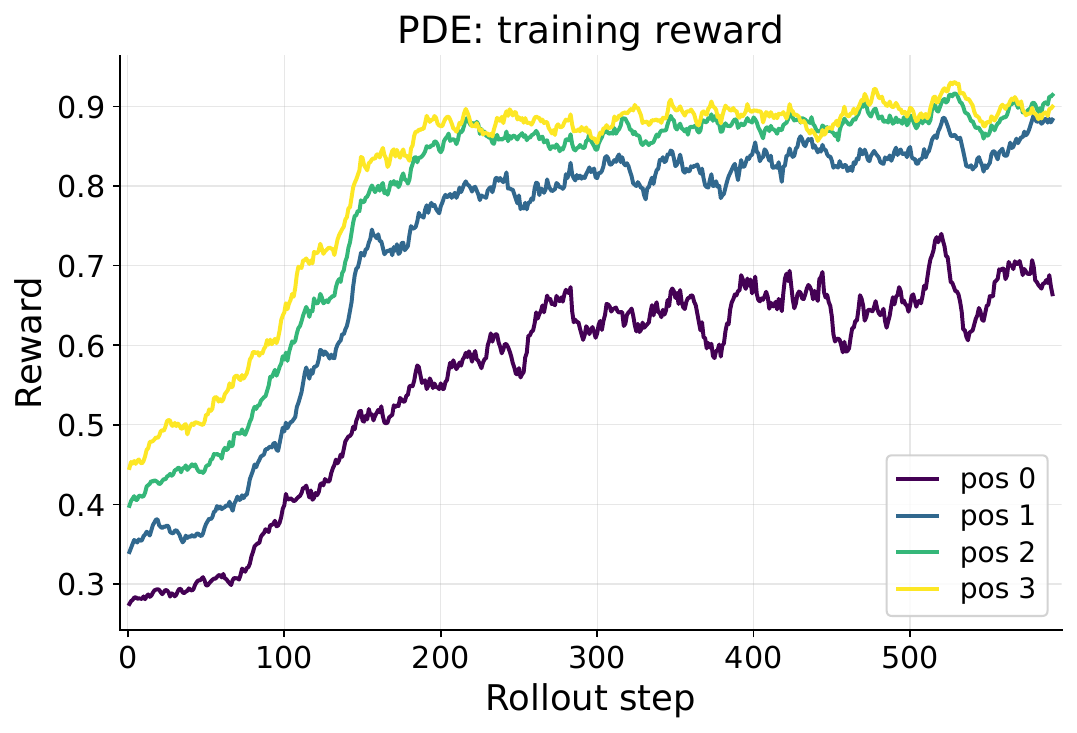}%
    \includegraphics[width=.45\textwidth]{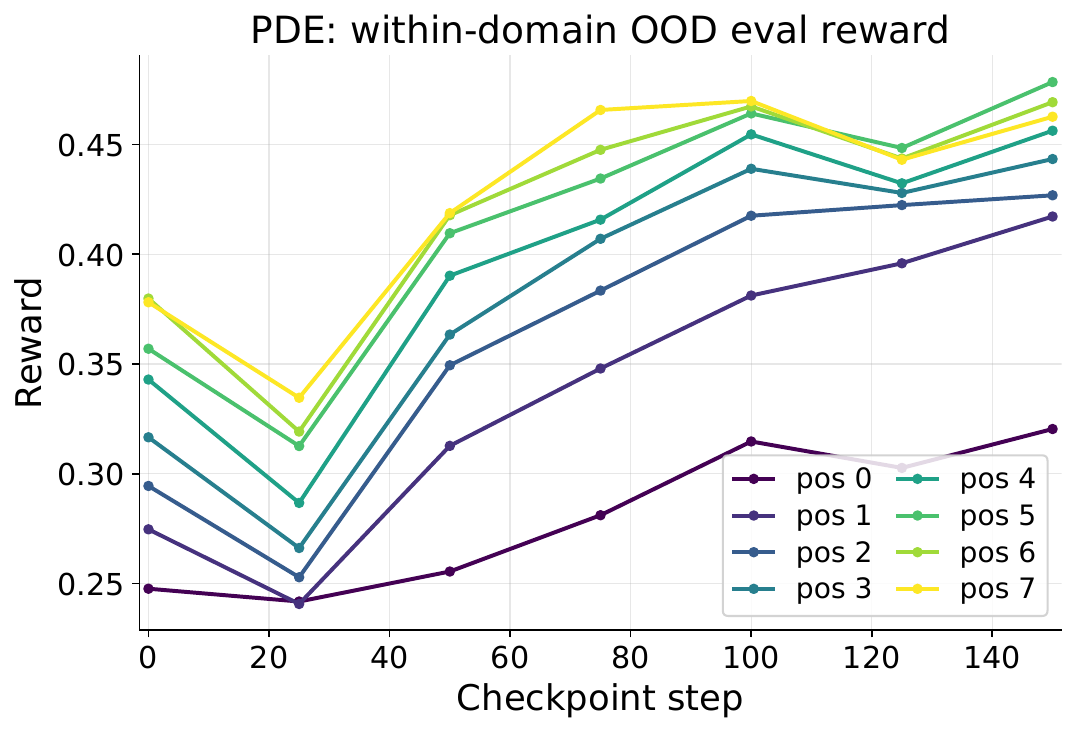}

    \includegraphics[width=.45\textwidth]{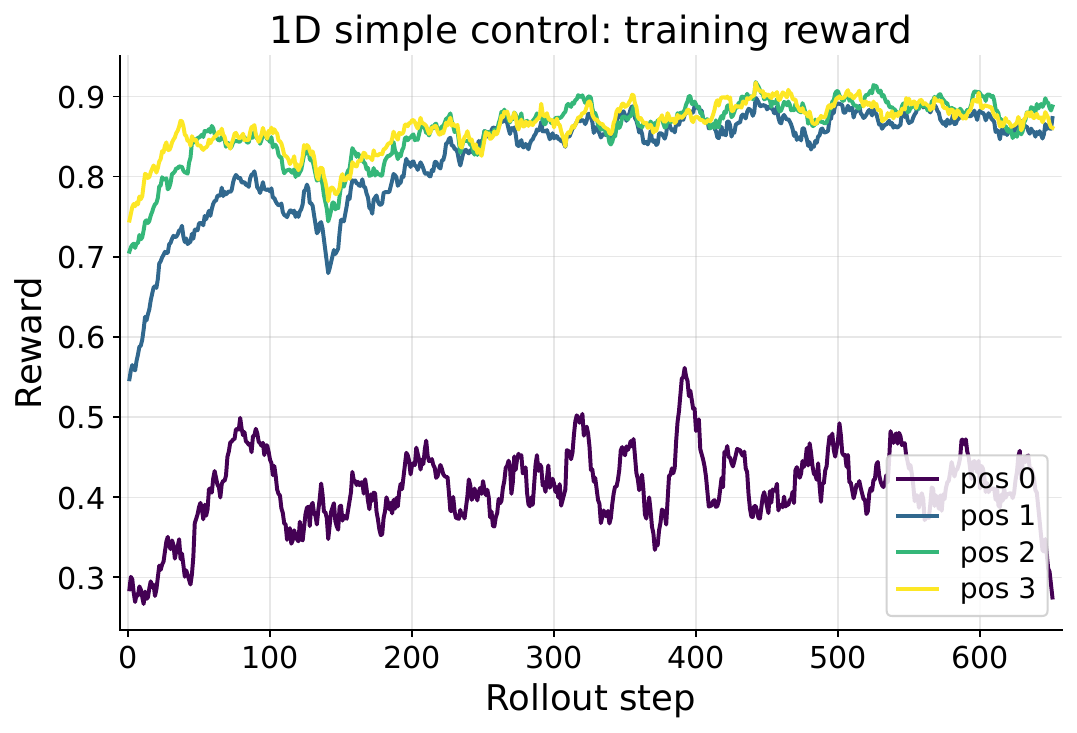}%
    \includegraphics[width=.45\textwidth]{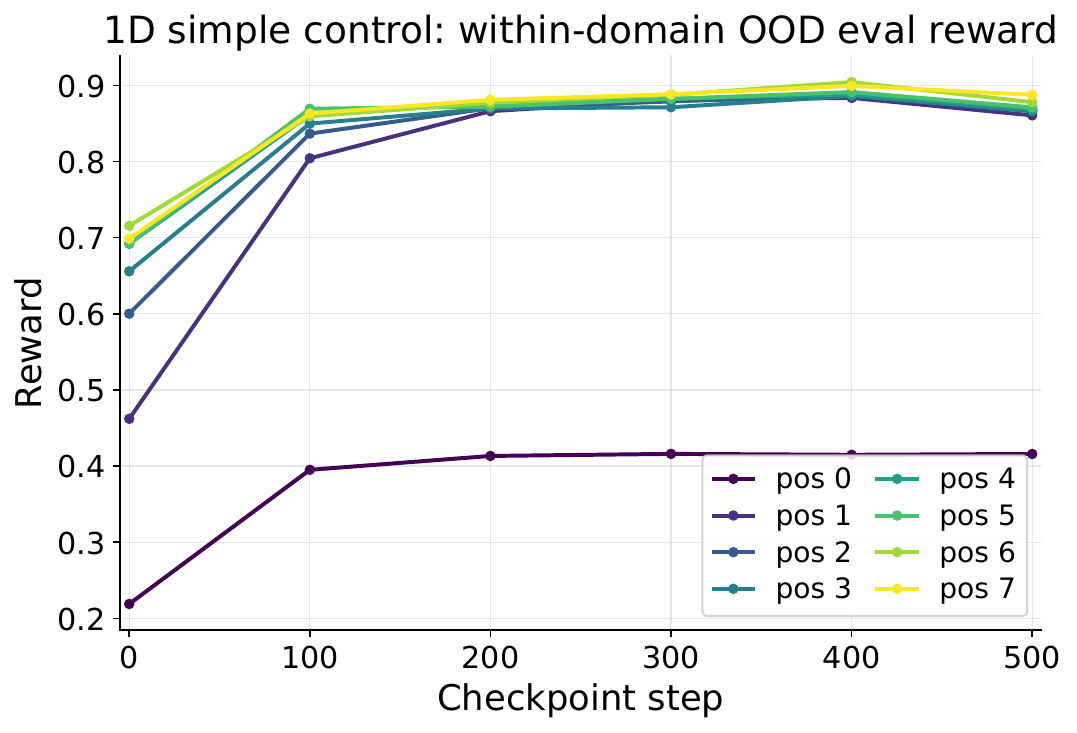}

    \includegraphics[width=.45\textwidth]{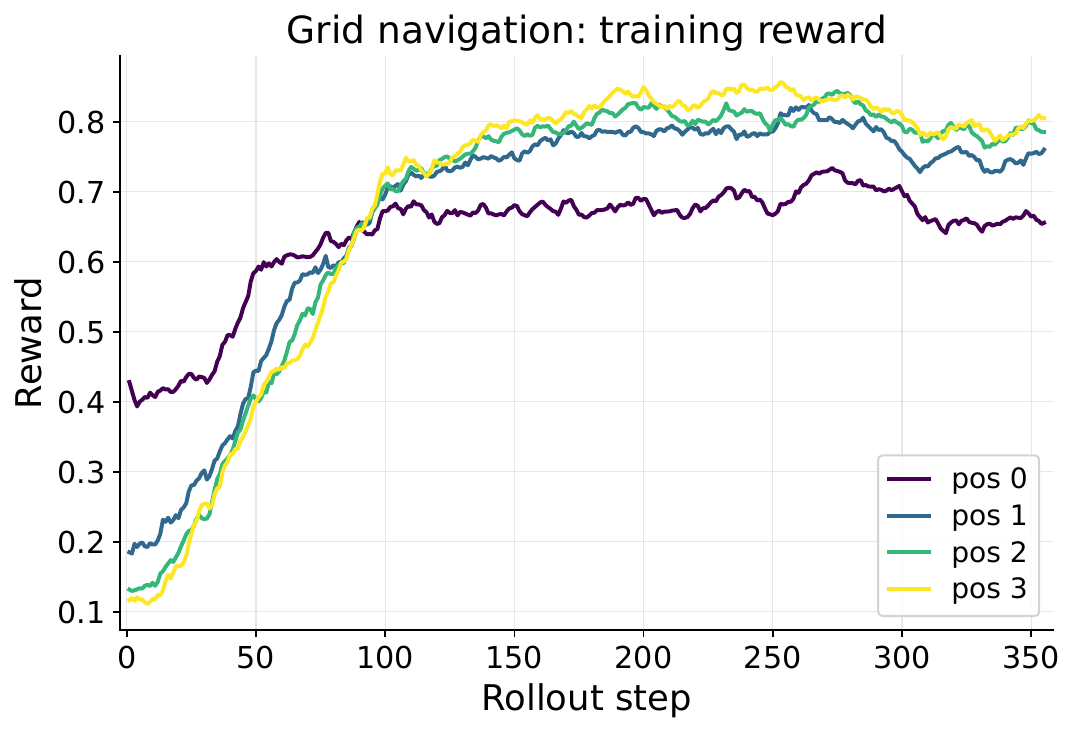}%
    \includegraphics[width=.45\textwidth]{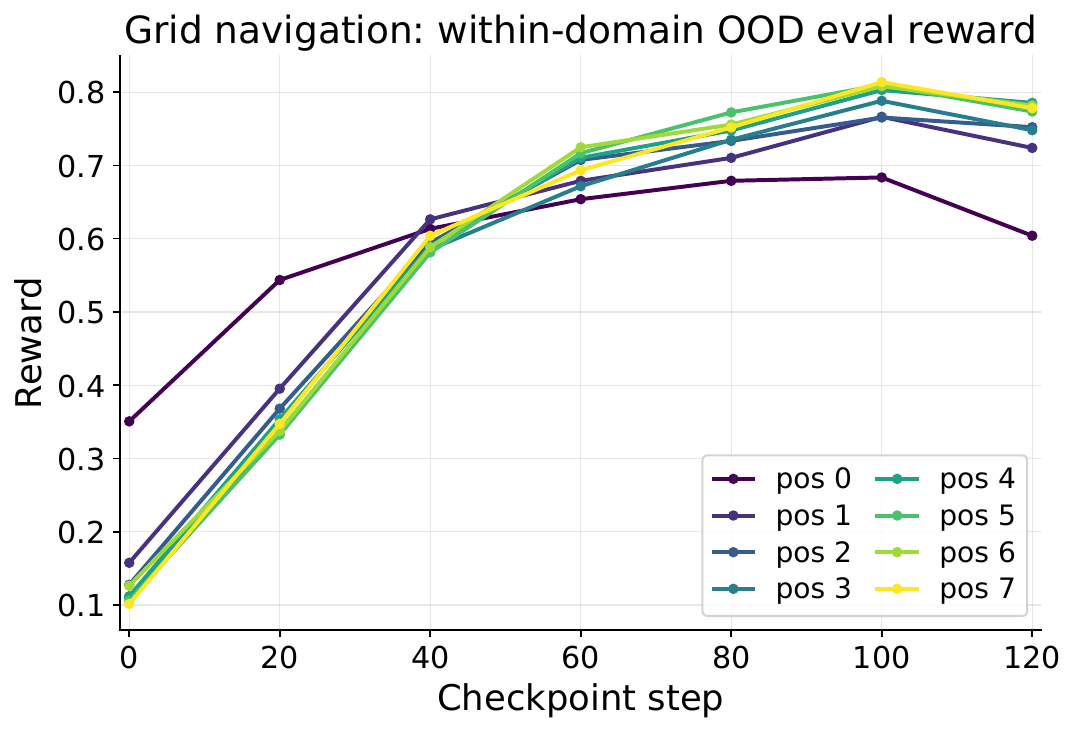}

    \caption{Empirical results for \qwensix and single-domain \codtrain by RL.
    Each row corresponds to one training trial for one domain.
    Training reward curves are smoothed with a running-average window of size 17.
    For evaluation curves, checkpoint step 0 corresponds to the initial \qwensix model.
    }
    \label{fig:exp_27b_single_domain_rl}
\end{figure}

\begin{figure}
    \centering

    \includegraphics[width=.45\textwidth]{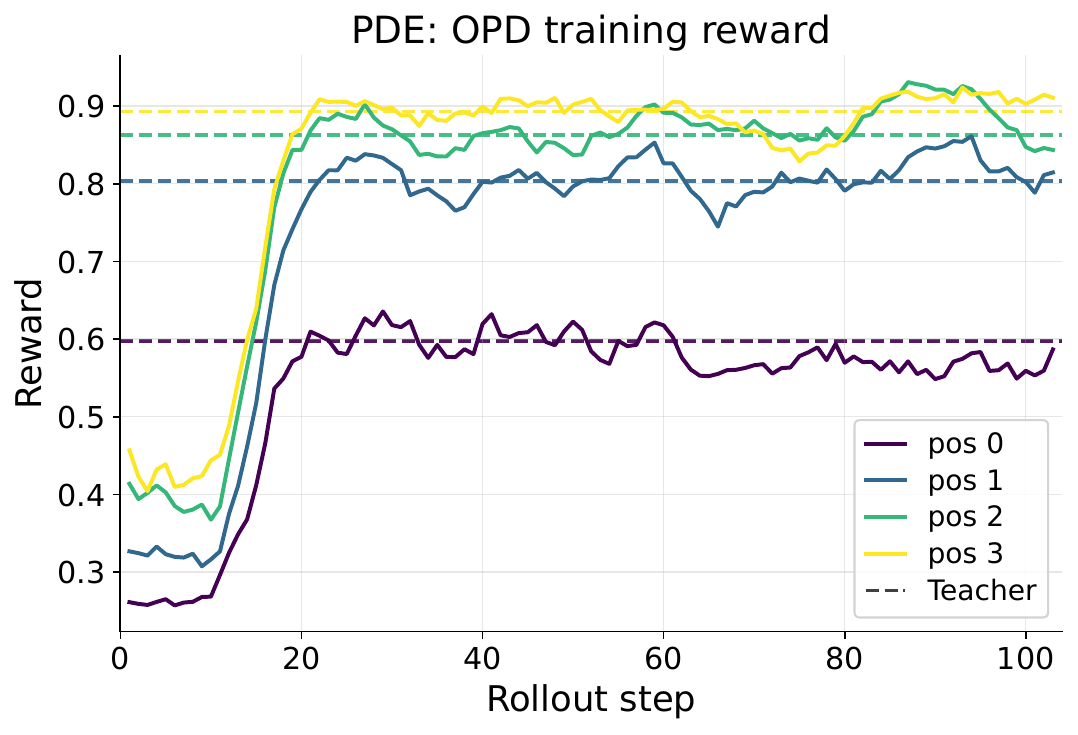}%
    \includegraphics[width=.45\textwidth]{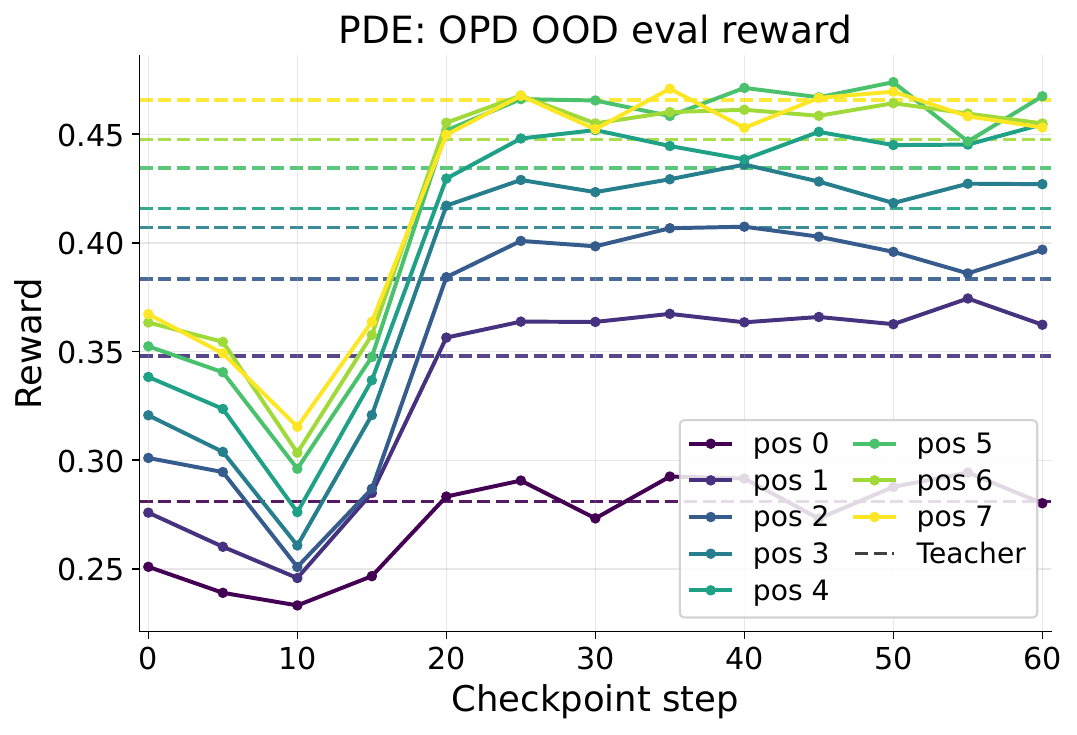}

    \includegraphics[width=.45\textwidth]{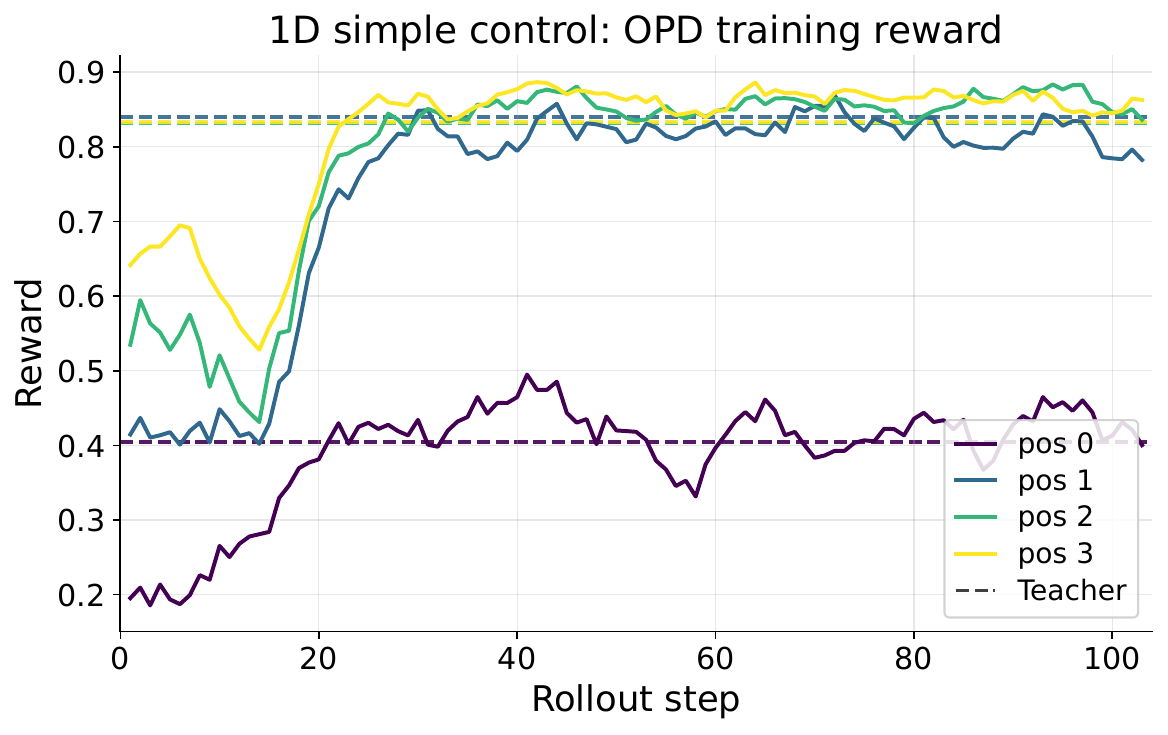}%
    \includegraphics[width=.45\textwidth]{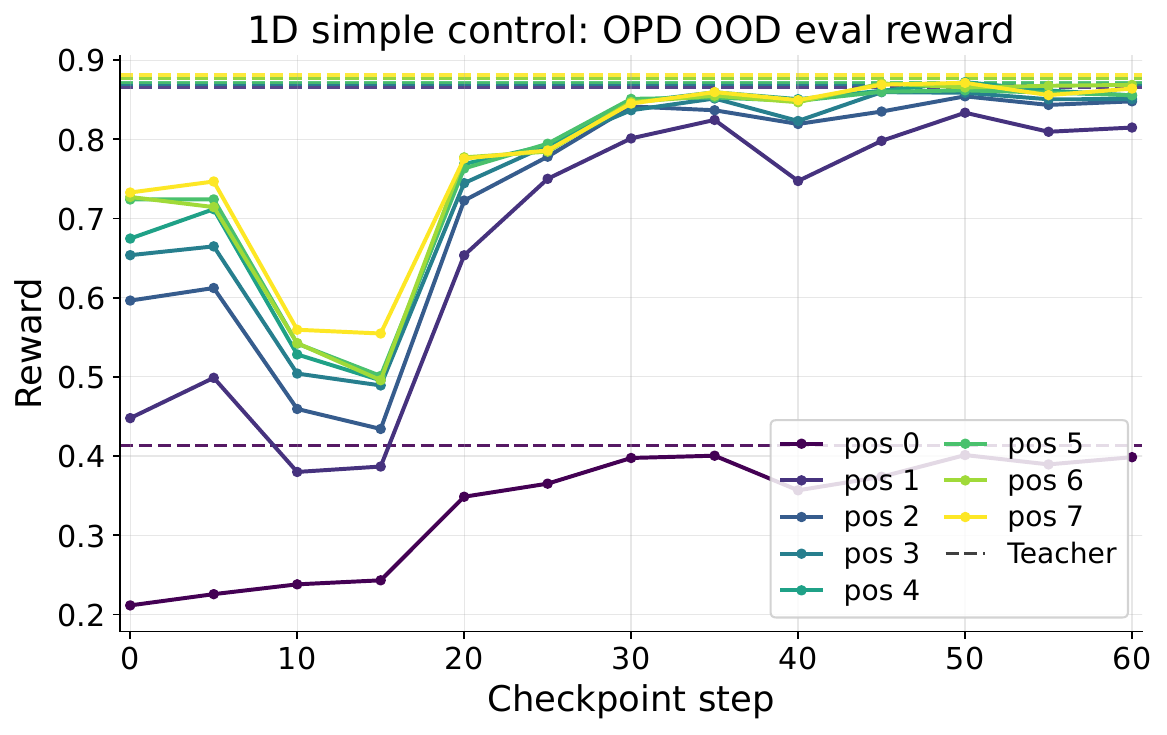}

    \includegraphics[width=.45\textwidth]{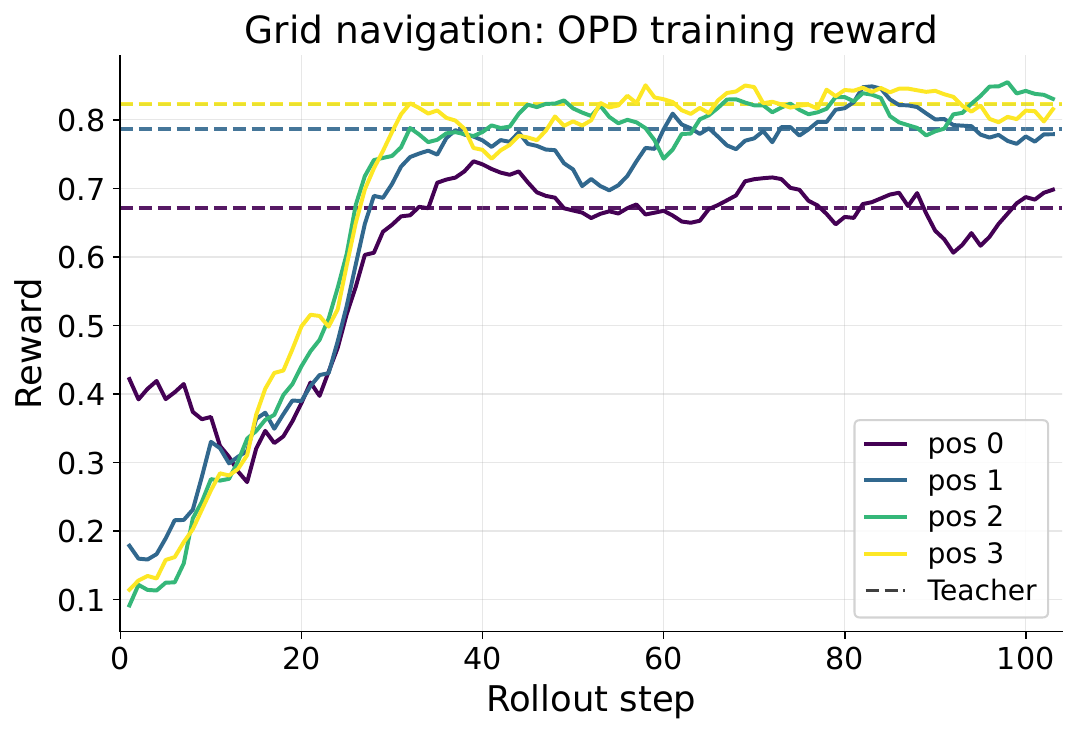}%
    \includegraphics[width=.45\textwidth]{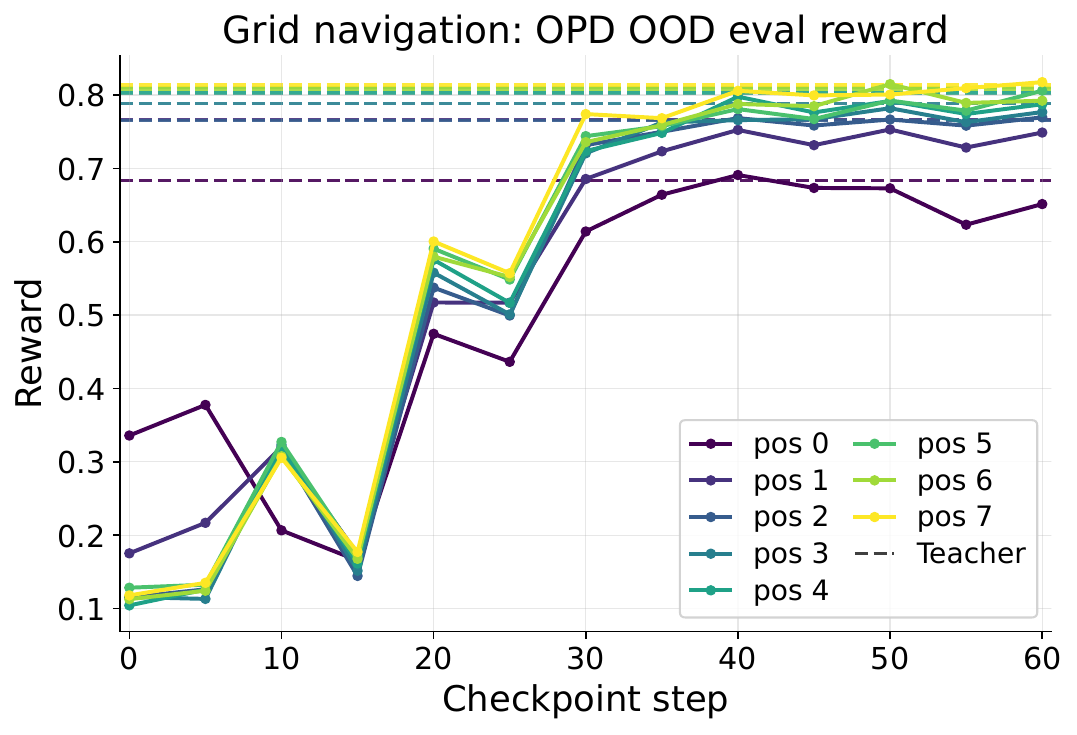}

    \caption{Empirical results for \qwensix and mixed-domain \codtrain by multi-teacher OPD.
    All plots in this figure correspond to a single OPD training trial with a mixture of three domains.
    Horizontal dashed lines represent the performance of teacher models, i.e., checkpoints from the single-domain RL experiments in Figure~\ref{fig:exp_27b_single_domain_rl}.
    Training reward curves are smoothed with a running-average window of size 7.
    For evaluation curves, checkpoint step 0 corresponds to the initial \qwensix model.
    }
    \label{fig:exp_27b_mixed_domain_opd}
\end{figure}

We further scale up our experiments to larger model sizes and more challenging environments:
starting from \qwensix, we aim for a 27B LLM that is proficient in all of the \pde, \control and \navigation domains.
The \codtrain process for achieving this goal consists of two stages: 
we first apply single-domain RL for each domain separately,
and then use multi-teacher OPD for merging the domain experts into a single model.

\paragraph{Stage 1: single-domain training by RL.}

For each domain, we apply \codtrain by RL to environments within a specific range of configurations, and evaluate the model checkpoints in out-of-distribution settings, using environments with different configurations and longer task sequences of length 8.
The results are shown in Figure~\ref{fig:exp_27b_single_domain_rl}: 
both training and evaluation reward curves grow with training steps,
and higher rewards can be achieved by the trained LLM at later positions within a sequence of tasks\footnote{
Careful readers might notice that in \navigation, the initial \qwensix model actually achieves higher rewards at position 0 than at later positions.
This indicates the limited capability of \qwensix in this domain, whose performance degrades when the partially revealed cost map and its self-generated hint are included in the prompt of a solve-task episode.}.

\paragraph{Stage 2: mixed-domain training by multi-teacher OPD.}

For each domain, we choose a teacher model from the previous RL stage by identifying the performance ceiling and picking the earliest model checkpoint that is sufficiently close to that ceiling (i.e., checkpoint step 75 / 200 / 100 for \pde / \control / \navigation, respectively), in order to reduce the risk of overfitting.
The multi-teacher OPD process is also initialized with \qwensix.
Empirical results of the OPD process can be found in Figure~\ref{fig:exp_27b_mixed_domain_opd}: both training and evaluation rewards in all three domains grow stably during the sample-efficient OPD process, eventually matching the performance of the teacher models.

\paragraph{Comparison with Qwen3.8 models.}

We compare the \cod-trained model (at checkpoint step 50 of the OPD process) with \qwensix, \qwentpeseven and \qwentpemax on the evaluation sets\footnote{We set \texttt{enable\_thinking = false} for Qwen3.6/3.8-27B and \texttt{reasoning\_effort = "low"} for \qwentpemax, to avoid LLM responses that are excessively and unnecessarily lengthy for the difficulty of tasks in our experiments.}.
The results in Figure~\ref{fig:exp_27b_compare_qwen38} show that our OPD checkpoint significantly outperforms Qwen3.6/3.8-27B and closely matches the performance of \qwentpemax in all three domains.

\section{Related works}
\label{sec:related_works}

\paragraph{Research and practice of ``lifelong agents''.}

There has been a surge of research on lifelong agents, which includes a dedicated workshop at ICLR 2026 \citep{lifelongagent}. 
On the theoretical side, recent work has argued that context management is crucial for boosting the computational expressivity of an LLM-based agent system \citep{cui2026positionturingcompletenessrealworldautoregressive}. 
On the practical side, there is growing interest in LLM-based AI agents with long-lifecycle deployment, in which an agent updates its own context --- e.g., in the form of persistent memory or skill sets --- and thereby self-improves its task-solving performance over time
\citep{hermesagent,openclaw,claudecode,qwenpaw,chhikara2025mem0buildingproductionreadyai}.

A notable gap in the literature is the lack of dedicated LLM post-training for the meta-capability required by a lifelong agent. As a result, practical AI systems often depend on human-crafted agent scaffolds with sophisticated hand-designed components. 
A related limitation is that frontier LLMs still fall short in environments with underspecification \citep{pu2026lhawcontrollableunderspecificationlonghorizon}, where agents must adapt online to a new environment, balance exploration and exploitation, and take actions conditioned on the context \citep{dou2026clbenchbenchmarkcontextlearning}. 
In contrast to these requirements, agents powered by current LLMs often act based on incorrect priors with over-confidence, or lose coherence after multiple rounds of interaction with the environment \citep{laban2026llms,wei2025groundedrealitylearningdeploying}.

An important design question in this area is what should be updated during learning: the context, the model weights, or both. Some approaches optimize context alone, such as prompts, few-shot examples, or skills \citep{zhao2024expelllmagentsexperiential,opsahl-ong-etal-2024-optimizing,wang2024voyager,agrawal2026gepa,ouyang2026reasoningbank,zhang2026agenticcontextengineeringevolving}, while others combine context optimization with updating model weights \citep{soylu-etal-2024-fine}. 
More recent work \citep{tiwari2026learningfastslowllms} integrates both perspectives, but trains the policy model for solving tasks via RL while relying on an external, fixed LLM to update the agent's states or context. 
Other recent work \citep{yi2026learningagentcompatiblecontextmanagement,li2026escapingcontextbottleneckactive} instead fixes the solve-task LLM and trains the context-management LLM via RL for long-horizon tasks. 
\cod offers an alternative way to integrate both components seamlessly: it trains model weights during \codtrain so that the LLM becomes proficient not only at solving tasks, but also at self-updating its own context about the environment during \coddeploy.

\paragraph{Meta reinforcement learning.}

\cod is closely related to meta reinforcement learning \citep{BOTVINICK2019408,stadie2019considerationslearningexploremetareinforcement,metarl-tutorial,han2026selfimproving}, especially the RL$^2$ paradigm \citep{duan2016rl2fastreinforcementlearning}. 
In the original RL$^2$ formulation, multiple solve-task episodes are concatenated into a single rollout trajectory, and the hidden state of a recurrent neural network (RNN) serves as the context transferred across episodes. 
One important difference is that, with an RNN, context is produced using a fixed amount of compute and has a fixed size; in contrast, with an LLM, context can have adaptive size and can be generated through a thinking process with adaptive compute, making it more computationally expressive \citep{merrill2024the,li2024chain,cui2026positionturingcompletenessrealworldautoregressive}. 
A second difference is that LLMs introduce new opportunities for out-of-distribution generalization, compared to RL in the pre-LLM era.

Several recent works attempt to adapt RL$^2$ to LLMs, including LaMer \citep{jiang2026metarl}, MAGE \citep{yang2026mage}, and Orbit \citep{lin2026scalingincontextonlinelearning}. These works differ from the present work in their problem settings and RL algorithms. 
In all of these works, the agent performs a sequence of episodes or attempts for the same task. This setting is closer to sequential test-time scaling (discussed later in this section) or a Ralph loop \citep{ralphloop}, rather than to the classic form of meta RL. In such settings, the sequence of episodes terminates upon success, whereas in \cod the agent must continue solving newly imposed tasks throughout deployment. 
LaMer \citep{jiang2026metarl} and MAGE \citep{yang2026mage} employ the GiGPO algorithm \citep{feng2026groupingroup}, which assumes that repeated environment states, called ``anchor states,'' can be identified across task-solving trajectories; \cod does not make such an assumption. 
Orbit \citep{lin2026scalingincontextonlinelearning} uses GRPO and treats the sum of rewards over a sequence of episodes as the final outcome reward. Such coarse-grained credit assignment inevitably limits the sequence lengths for which effective training is possible. In our early experiments, this approach was not feasible for effective training on \flobs, whereas our fine-grained credit assignment enables effective training across a broad range of settings.

\paragraph{LLM inference scaling.}

\coddeploy is also related to LLM test-time scaling. 
In this setting, an agent attempts to recursively self-improve its performance \citep{recursiveselfimprovement} through repeated LLM calls, which may be organized sequentially, in parallel, or through a combination of both \citep{lightman2024lets,snell2025scaling,chen2026provable}. Practical examples include the Ralph loop \citep{ralphloop} and auto-research systems \citep{novikov2025alphaevolvecodingagentscientific}. 
The connection is straightforward: these settings can be viewed as a special case of \coddeploy in which the task sequence $x_0, x_1, x_2, \dots$ happens to consist of repeated instances of exactly the same task. 
Recent works have also proposed RL training methods specifically aimed at improving the capability for test-time inference scaling \citep{hu2026pacorelearningscaletesttime,kimiteam2026kimik25visualagentic}.

Inference scaling can also be incorporated into the rollout stage of RL. Examples include reflection-and-retry methods \citep{shi2026r3lreflectthenretryreinforcementlearning,song2026expandingcapabilitiesreinforcementlearning,shinn2023reflexionlanguageagentsverbal} and approaches that incorporate skills during rollout \citep{lu2026skill0incontextagenticreinforcement}. 
The objective of inference scaling in these methods is to generate rollout trajectories of higher quality. However, the policy model is still trained to solve tasks from scratch, as the additional context (e.g., reflection or skills) is removed when rollout trajectories are converted into training samples for gradient computation. By contrast, \cod explicitly trains the LLM to adapt to a new environment at deployment, and solve new tasks conditioned on its continuously updated context.

\section{Discussions}
\label{sec:discussions}

\paragraph{Limitations and future work.}

This work has introduced a general \cod framework,
provided proof-of-concept implementations of it,
and presented promising empirical results.
Still, many research questions remain open and call for further works.
For example, there remain some heuristic augmentations and potential caveats in our current GRPO-style RL algorithm (cf.~Section~\ref{subsec:rl_algorithm_infrastructure} and Appendix~\ref{subsec:complete_rl_algorithm}); 
future works might aim for a more principled and theoretically grounded RL algorithm with better empirical performance.
Moreover, future works might continuously {scale up} \cod along various dimensions, for example:
\begin{itemize}
\item Designing more diverse and challenging environments (tailored to concurrent frontier LLMs) for training and evaluating the \cod meta-capability;
\item Validating \cod in broader settings, e.g.,
efficacy with longer task sequences in potentially non-stationary environments,
OOD generalization across a wider range of environments and domains,
or evaluation with recent benchmarks relevant to continual learning, loop engineering, and recursive self-improvement \citep{mittal2026latentgymtestbedcrosstaskexperiential,zhu2026edgebenchunveilingscalinglaws,asawa2026continuallearningbenchevaluating,harrington2026doescontinuallearningrequire,shu2026agentclrigorousevaluationcontinual,wang2026evopolicygymevaluatingautonomouspolicy,chi2026frontierengbenchmarkingselfevolvingagents,zheng2025lifelongagentbenchevaluatingllmagents,wei2026evomemorybenchmarkingllmagent,he2026memoryarenabenchmarkingagentmemory,jiang2026seaevalbenchmarkevaluatingselfevolving,zhang2026skillflowbenchmarkinglifelongskilldiscovery,battleday2026digbenchdiscoverygames};
\item Extending the \cod framework and implementations to more powerful context-management mechanisms, or more flexible rollout patterns than those visualized in Figure~\ref{fig:cod-viz}.
\end{itemize}

\paragraph{Integration into LLM post-training pipelines?}

It is worth noting that \cod is complementary (rather than alternative) to the standard task-by-task RL paradigm that aims for enhancing domain-specific capabilities,
akin to the relation between fluid and crystallized intelligence \citep{chollet2019measureintelligence}.
A natural question is thus about how to integrate \cod into existing LLM post-training pipelines.
One straightforward idea is to regard \codtrain as an additional sequential stage, or mix up a \codtrain objective with the original RL objective.
Another potential idea is to first train a \cod teacher model by \codtrain, and then use it for model merging --- together with other domain-specific teacher models --- via on-policy distillation \citep{lu2025onpolicydistillation,agarwal2024onpolicy,yang2025qwen3technicalreport,qwen35blog,deepseek-v4,glm5team2026glm5vibecodingagentic}.
We are excited to see what \cod can potentially bring to the next generation of frontier LLMs.

\section*{Acknowledgements}

We would like to thank Yuchang Sun for helpful discussion and feedback, as well as
Xuchen Pan and Yushuo Chen for infrastructure support related to \trinity.

\bibliographystyle{plainnat}
\bibliography{references}

\newpage
\appendix

\section{The complete RL algorithm}
\label{subsec:complete_rl_algorithm}

\paragraph{Notations.}

We denote a sequence of $S$ tasks by $x_0, x_1, \dots, x_{S-1}$,
and assume access to a group of $G$ end-to-end rollout trajectories for each task sequence during \codtrain.
Let $\rx_{i,j}$  (resp.~$\rz_{i,j}$) denote the episode-wise reward for the $j$-th episode solving $x_j$ (resp.~updating $z_j$) within the $i$-th trajectory, where $1 \le i \le G$ and $0 \le j \le S - 1$.

\paragraph{Credit assignment and advantage calculation.}

Our method is visualized in Figure~\ref{fig:viz_cod_advantage} and formalized below.
We define the return for each episode --- indexed by $i$ and $j$ --- as
\begin{align*}
\Rx_{i,j} \coloneqq \frac{1}{S - j} \sum_{j \le \ell \le S-1} \rx_{i, \ell}
\quad \text{and} \quad
\Rz_{i,j} \coloneqq \frac{1}{S - j + 1} \Big( \rz_{i,j} + \sum_{j \le \ell \le S-1} \rx_{i, \ell} \Big).
\end{align*}
Then, the baseline at position $j$ is defined as
\begin{align*}
\Rbarx_{j} \coloneqq \frac{1}{G} \sum_{1 \le i \le G} \Rx_{i,j}
\quad \text{and} \quad
\Rbarz_{j} \coloneqq \frac{1}{G} \sum_{1 \le i \le G} \Rz_{i,j}.
\end{align*}
Finally, we define advantages as
\begin{align*}
\Ax_{i,j} \coloneqq \Rx_{i,j} - \Rbarx_{j}
\quad \text{and} \quad
\Az_{i,j} \coloneqq \Rz_{i,j} - \Rbarz_{j}.
\end{align*}

\begin{remark}
Some recent works \citep{li2026rubricemmetarlrubricguidedpolicy,xiao2026metareinforcementlearningselfreflectionagentic} adopt credit-assignment methods that are similar to ours, albeit in different RL settings.
\end{remark}

\paragraph{Gradient calculation.}

We adopt token-mean aggregation for the loss and gradient terms within one batch of training samples.
Consider a batch with $N$ tokens in total; suppose that the $i$-th token has advantage $A_i$, probability by its corresponding rollout policy $\pioldi$, and probability by the current policy model $\pithetai$, where $\btheta$ is the model parameters under training.

Standard PPO \citep{schulman2017proximal} and GRPO \citep{deepseek-math} calculate the gradient update as
\begin{align}
    \bg \coloneqq \frac{1}{N} \sum_{1 \le i \le N} \nablatheta \log \pithetai \cdot \Ai \cdot \frac{\pithetai}{\pioldi} \cdot \Mi,
    \label{eq:gradient_update_grpo}
\end{align}
where $\pithetai / \pioldi$ is the importance-sampling (IS) weight, and $\Mi$ denotes a one-side clipping mask, specified by hyperparameters $\epshigh$ and $\epslow$:
\begin{align*}
    \Mi \coloneqq  \ind\Big( \Ai > 0, \frac{\pithetai}{\pioldi} \le 1 + \epshigh \Big) + \ind\Big( \Ai < 0, \frac{\pithetai}{\pioldi} \ge 1 - \epslow \Big)
\end{align*}
\cite{yao2026grouprelative} investigate the REINFORCE-with-clipping (\rec) series, and in particular, a \reconesidenois variant that discards the (potentially unstable) IS weights:
\begin{align}
    \bg \coloneqq \frac{1}{N} \sum_{1 \le i \le N} \nablatheta \log \pithetai \cdot \Ai \cdot \Mi.
    \label{eq:gradient_update_recnois}
\end{align}
\reconesidenois has been theoretically justified and shown empirical advantages over standard GRPO (with IS weights) in the outcome-reward RLVR setting \citep{yao2026grouprelative}.
However, we have observed in our early experiments that, when adapted to the \codtrain setting, this algorithm could incur training instability on the \alchrand domain.
In particular, such instability appears to correlate with the decay of the mean-advantage metric $\sum_{i} \Ai / N$, which starts from a negative value and keeps decreasing during \codtrain.
We suspect that this causes an imbalance between positive and negative gradients, leading to training instability.
This motivates us to apply a token-level re-weighting heuristic to the batch if its mean advantage is negative, in which case the gradient update becomes
\begin{align}
    \bg \coloneqq \frac{1}{N} \sum_{1 \le i \le N} \nablatheta \log \pithetai \cdot \Ai \cdot \exp\Big(\frac{\Ai}{T}\Big) \cdot \Mi.
    \label{eq:gradient_update_reweight}
\end{align}
Here, the temperature parameter $T > 0$ is set adaptively for each batch: 
we select $T$ (via bisection) such that $\sum_{i} \Ai \exp(\Ai / T) / N \approx 0$,
although we also impose a hard constraint $T > 0.8$ to avoid numerical instability caused by a small $T$.
This re-weighting mechanism is analogous to the \redweight method investigated by \cite{yao2026grouprelative}, albeit with different rationales and purposes.

\paragraph{Experiments.}

We conduct empirical comparisons between three methods of calculating the gradient update, when running \codtrain on \alchrand environments:
Eq.~\eqref{eq:gradient_update_grpo}, the vanilla GRPO method;
Eq.~\eqref{eq:gradient_update_recnois}, the \reconesidenois method that discards the IS weights in GRPO; and
Eq.~\eqref{eq:gradient_update_reweight}, our final method that augments \reconesidenois with an adaptive re-weighting mechanism.

\begin{figure}
    \centering
    \includegraphics[width=.4\textwidth]{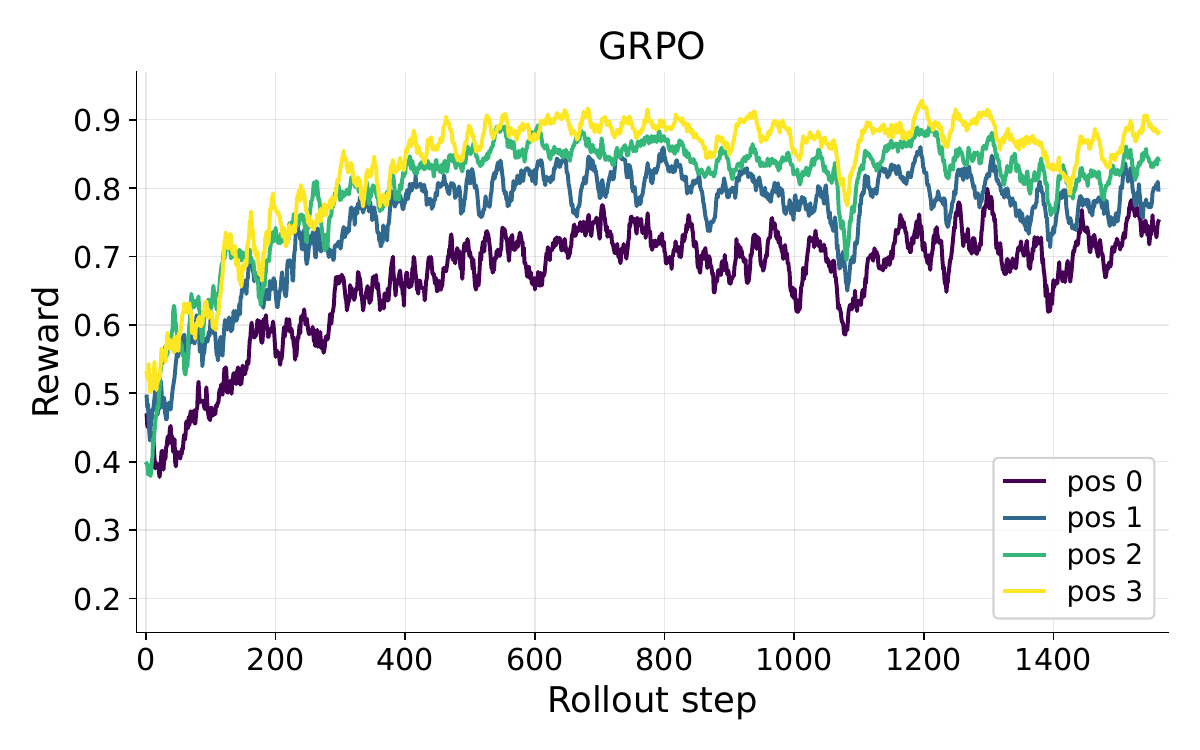}
    \includegraphics[width=.4\textwidth]{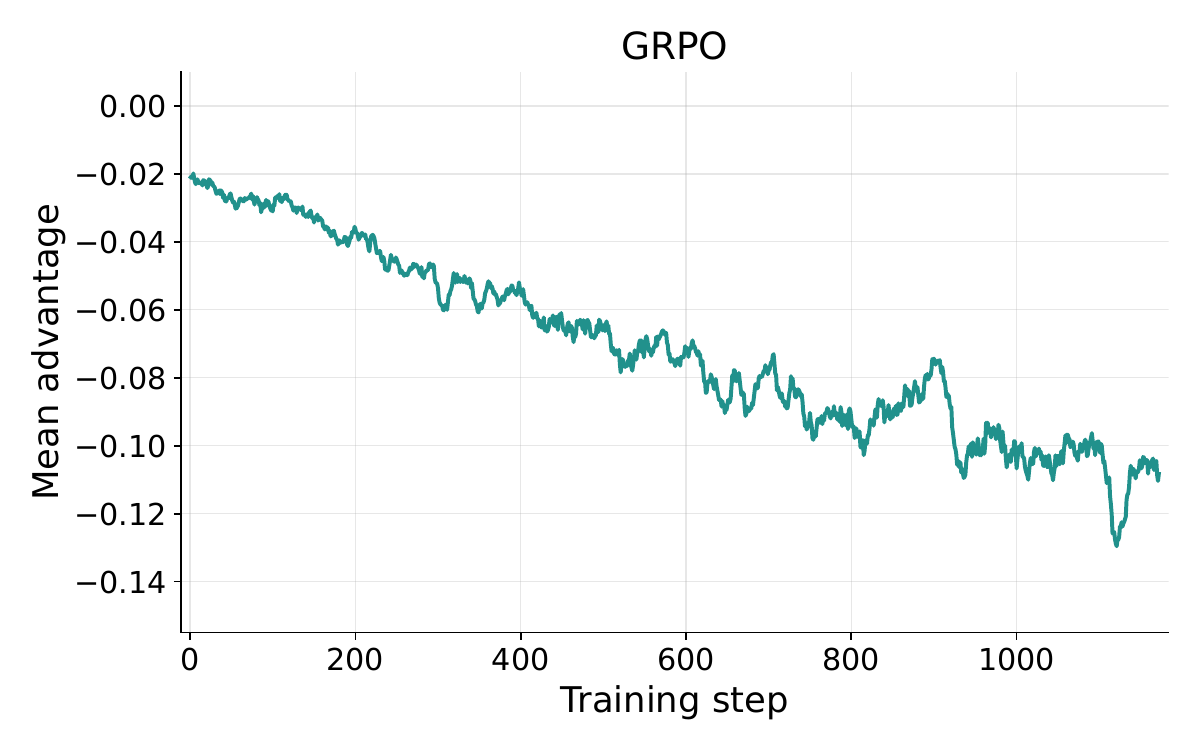}
    \includegraphics[width=.4\textwidth]{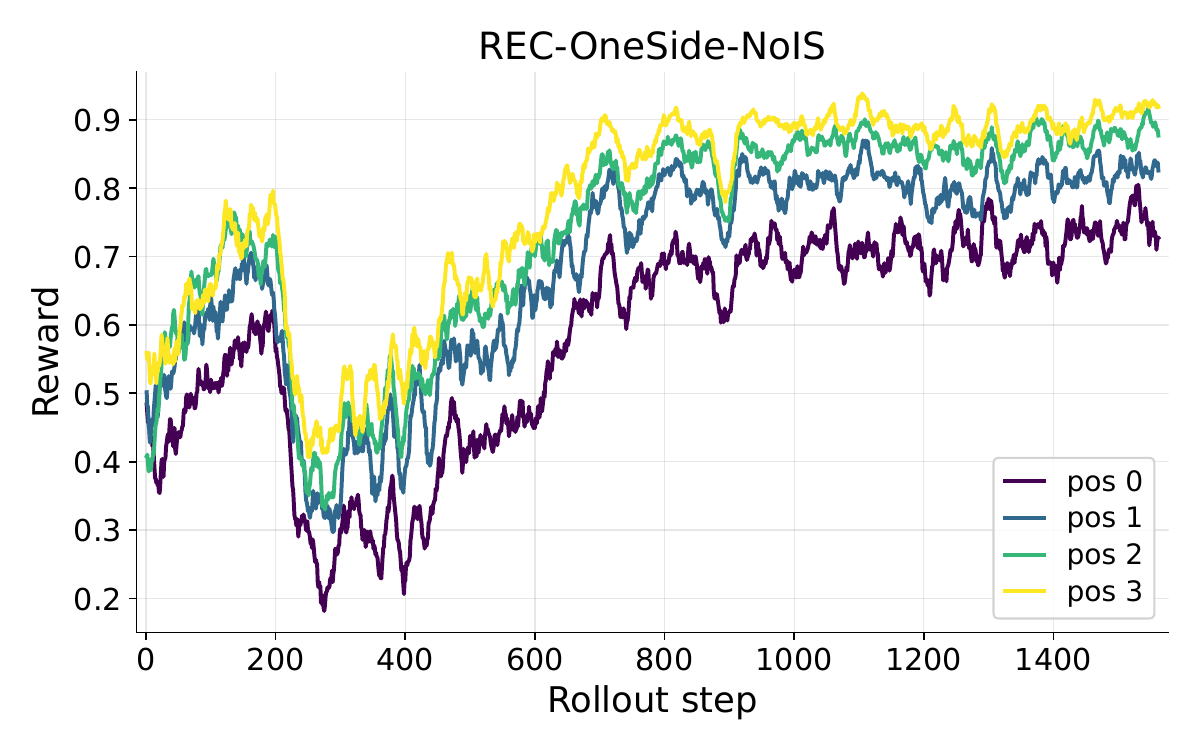}
    \includegraphics[width=.4\textwidth]{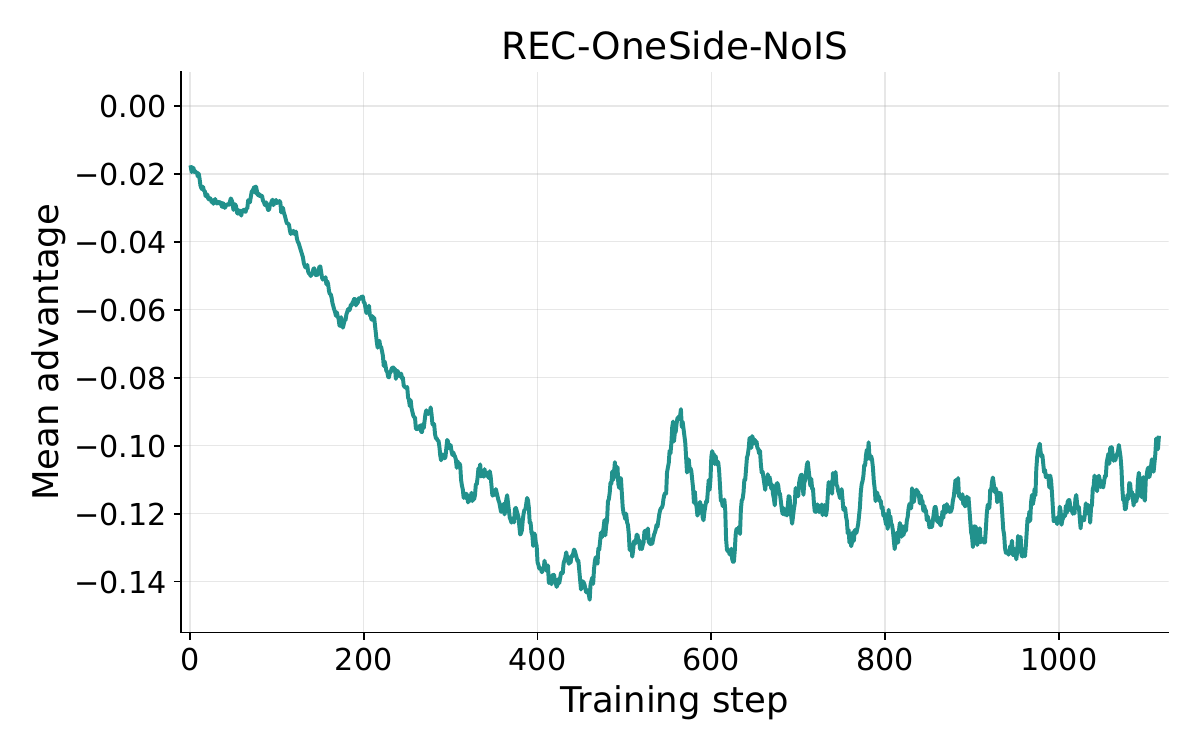}
    \includegraphics[width=.4\textwidth]{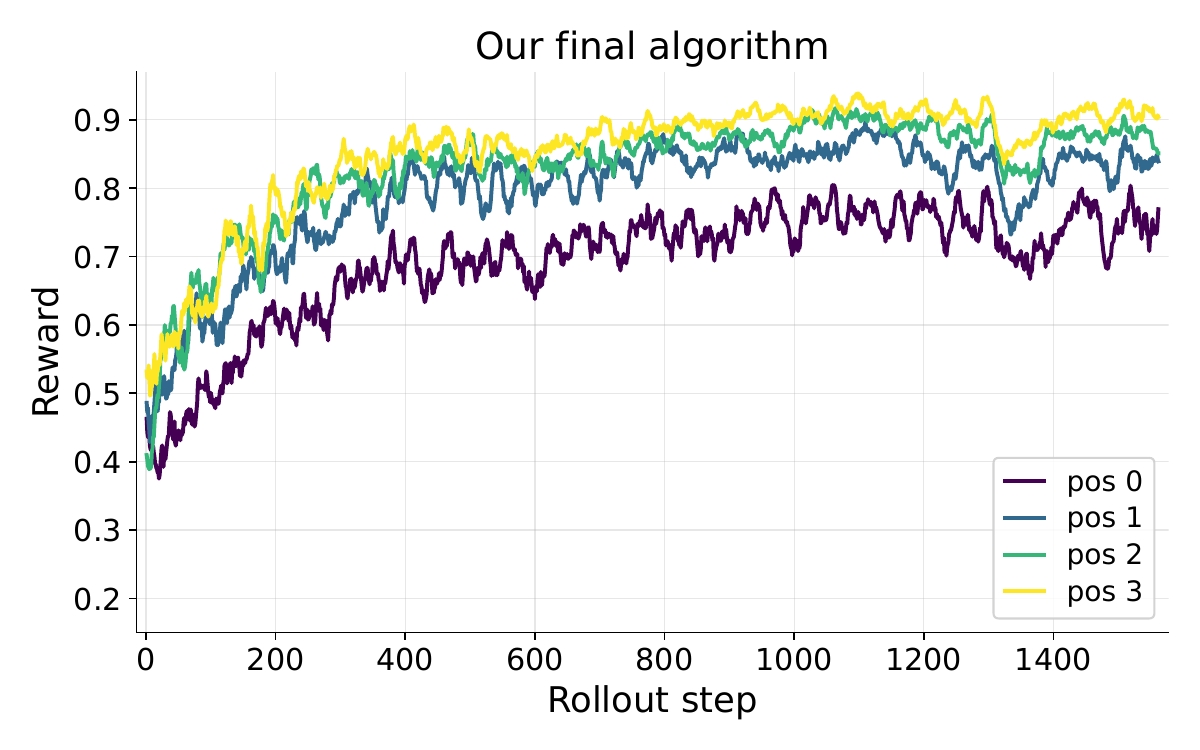}
    \includegraphics[width=.4\textwidth]{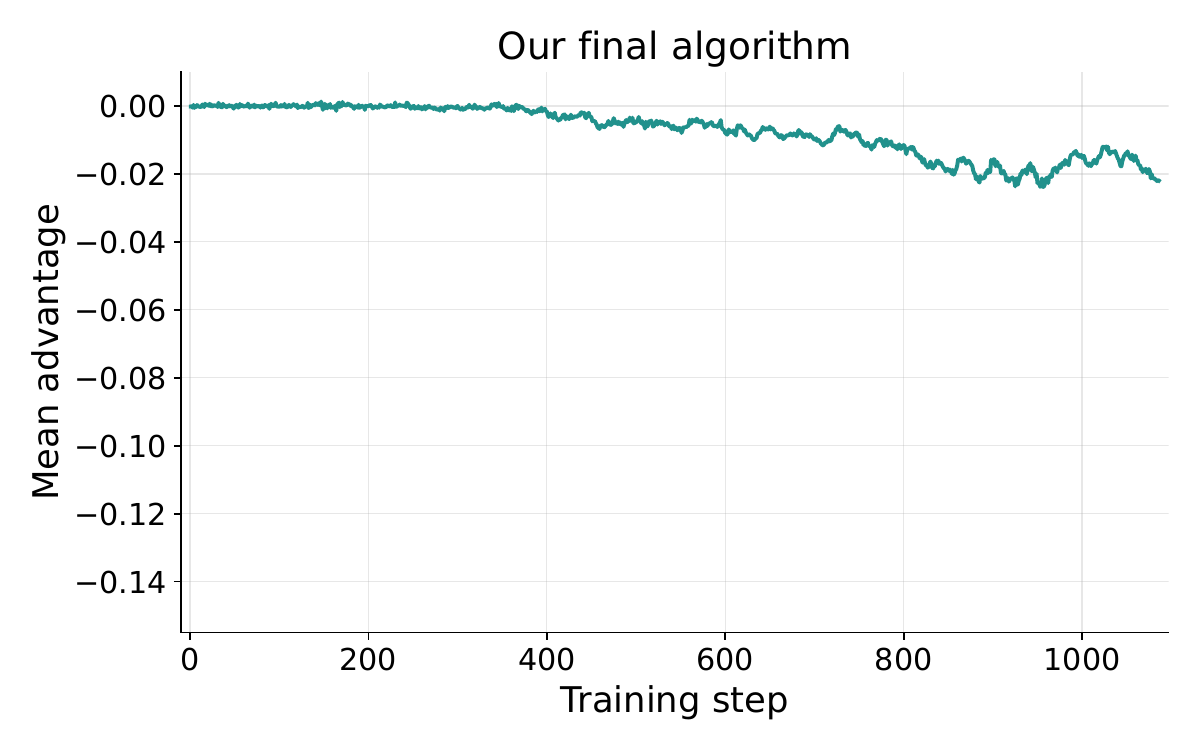}
    \caption{Empirical comparison between three RL algorithms.
    All curves are smoothed with a running-average window of size 17.
    }
    \label{fig:exp_compare_rl_algorithms}
\end{figure}

The results are shown in Figure~\ref{fig:exp_compare_rl_algorithms}.
The vanilla GRPO method encounters unstable reward curves and a decaying mean advantage,
and the issues are even more severe for \reconesidenois.
In comparison, our final algorithm leads to the most stable training process among the three methods.

\paragraph{Remarks.}

Although our final RL algorithm works reasonably well in our experiments,
there is still room for improvements.
(1) We use the average of rewards-to-go as the return for each episode, which might become problematic as the length of task sequences scales up.
In this case, it could be beneficial to incorporate discounting or a finite sliding window for calculating returns.
(2)
Another potential caveat in our baseline calculation is that,
episodes at the same position from multiple rollout trajectories are not fully comparable.
In particular, episodes from different rollout trajectories for solving the same task $x_i$ start with different context $z_i$;
update-context episodes at the same position also face a similar issue.
While choices of the baseline do not affect the expectation of the policy gradient estimate (according to classic RL theory), they do make differences in practical finite-sample settings.
Future work might aim for a more principled and theoretically grounded RL algorithm tailored to \codtrain.

\section{Hyperparameters and configurations}
\label{subsec:exp_config}

Table~\ref{tab:draft_hyperparams_shared} lists the hyperparameters for \cod training and evaluation in our experiments. 
Table~\ref{tab:draft_env_config} summarizes the configurations of tasks and environments.
For brevity, here we only present the hyperparameters and configurations relevant to our \qweneight experiments.
For the \qwensix counterparts, we refer interested readers to our released codebase.

\begin{table}[ht]
  \centering
  \caption{Hyperparameters for \cod training and evaluation.}
  \label{tab:draft_hyperparams_shared}
  \renewcommand{\arraystretch}{1.05}
  \begin{tabular}{ll}
    \toprule
    Parameter & Value \\
    \midrule
    Initial model                      & \qweneight (thinking disabled) \\
    Task sequence length, train / eval & 4 / 8 \\
    \midrule
    \multicolumn{2}{c}{Optimization} \\
    Optimizer                       & AdamW ($\beta_1 = 0.9,\ \beta_2 = 0.999$) \\
    Learning rate                   & $1.0 \times 10^{-6}$ \\
    Gradient clipping               & 1.0 \\
    KL coefficient                  & 0 \\
    Training epochs                 & 1 \\
    Max staleness                   & 20 \\
    Synchronization interval        & 4 \\
    \midrule
    \multicolumn{2}{c}{Rollout and sampling} \\
    Rollout batch size              & 32 \\
    Rollout group size              & 8 \\
    Sampling temperature            & 1.0 \\
    Max prompt tokens, train / eval & $10{,}000$ / $30{,}000$ \\
    Max response tokens (hard cap)  & $2{,}000$ \\
    Max sequence tokens, train / eval  & $12{,}000$ / $32{,}000$ \\
    Suggested (by prompt) response length, train / eval & $1{,}000$ / $1{,}500$ \\
    Within-episode context-compression mode        & keep-all \\
    \midrule
    \multicolumn{2}{c}{RL algorithm} \\
    Policy loss             & \reconesidenois, $\epslow = \epshigh = 0.2$ \\
    Return estimator                & mean of rewards-to-go \\
    Advantage baseline              & per-position group mean \\
    Advantage normalization         & none \\
    Adaptive \redweight             & bisection with constraint $T > 0.8$ \\
    \midrule
    \multicolumn{2}{c}{Reward shaping} \\
    Hint length-penalty coefficient       & 0.1 \\
    Solve-task length-penalty coefficient & 0.1 \\
    \bottomrule
  \end{tabular}
\end{table}

\begin{table}[ht]
  \centering
  \caption{Configurations of tasks and environments. A value spanning both columns is shared by the easy and hard variants. \terminal is used only for cross-domain evaluation, with a single configuration. Each dataset has $50{,}000$ training instances and $4{,}000$ test instances.}
  \label{tab:draft_env_config}
  \renewcommand{\arraystretch}{1.1}
  \begin{tabular}{l cc}
    \toprule
    Parameter & Easy & Hard \\
    \midrule
    \multicolumn{3}{c}{\flobs} \\
    Map side length          & $4$ to $5$ & $6$ to $7$ \\
    Frozen-tile probability  & \multicolumn{2}{c}{$0.6$ to $0.7$} \\
    Environment step limit   & 6 & 8 \\
    Agent action budget      & 8 & 10 \\
    \midrule
    \multicolumn{3}{c}{\alchrand} \\
    Number of tiers          & \multicolumn{2}{c}{$3$ to $4$} \\
    Number of base elements  & $3$ to $4$ & $4$ to $6$ \\
    Recipes per element      & 1 & 1 to 2 \\
    Round limit              & 8 & 15 \\
    Material mode            & unlimited & limited \\
    \midrule
    \multicolumn{3}{c}{\terminal} \\
    Agent action budget      & \multicolumn{2}{c}{12} \\
    Composite-task ratio     & \multicolumn{2}{c}{0.5} \\
    \bottomrule
  \end{tabular}
\end{table}

\FloatBarrier

\clearpage
\newpage

\section{Prompt templates}
\label{subsec:prompt_templates}

Tables~\ref{tab:draft_prompt_update_context_sys}, \ref{tab:draft_prompt_frozenlake_obscure_sys}, \ref{tab:draft_prompt_alchemy_random_sys} and~\ref{tab:draft_prompt_terminal_sys} present the prompt templates used in our \qweneight experiments. 
Fields in double braces are filled in at runtime.
We refer interested readers to the released codebase for prompt templates relevant to our \qwensix experiments.

\makeatletter
\let\c@lstlisting\c@table
\makeatother
\renewcommand{\lstlistingname}{Table}
\lstset{captionpos=t, abovecaptionskip=2pt, belowcaptionskip=4pt}

\begin{lstlisting}[frame=tb, caption={Prompts for the update-context episode, which are generic and applicable to diverse environments.}, label={tab:draft_prompt_update_context_sys}]
You are a helpful assistant and a cross-task learning agent. Your responsibility is to solve **a sequence of different-but-related tasks** within the same environment, while **transferring informative hints across tasks** that can help achieve better task-solving performance.

## Problem setting

**In the current session, your job is to update the hints.**

To be concrete, the user provides:
1. Previous hints that were used to guide task solving;
2. A new task-solving trajectory.

Based on this information, you need to generate updated hints that can assist in solving future tasks within the same environment more effectively, by
- Preserving useful guidance and information from the previous hints, while discarding incorrect ones;
- Incorporating new lessons and information that can be learned from the new trajectory.

## General principles

- You are in an environment that might be stationary or non-stationary, and relation between tasks in this environment is initially unknown. One general principle is thus to generate **informative** hints that can serve as useful priors when solving a new task. If you are uncertain about the correctness or usefulness of certain hints, feel free to include them in your response and briefly mention uncertainty.
- There are at least two major categories of useful hints: (1) revealed information / clues about the environment that were initially unknown; (2) task-solving techniques with validated efficacy. With that said, your updated hints can certainly go beyond these two categories.

## Response format

First think about what hints to return, then provide your final answer. Format your complete response as follows:
```
[THINKING]
--- Start of updated hints ---
[HINTS]
--- End of updated hints ---
```
where [THINKING] and [HINTS] should be replaced with your actual thinking process and updated hints, respectively.

Other requirements:
- Keep your thinking process concise. Avoid overthinking.
- Keep your updated hints concise if possible, containing the most critical and helpful information. Avoid generic and uninformative statements. Avoid repeating the system prompt within the task-solving trajectory.
- Make sure that your updated hints can be potentially helpful for other related tasks, rather than specific to the task in the provided trajectory.
- Use standard Markdown format for your updated hints. Simpler structures (e.g., bullet points) are preferred. If you do need to use sections, you should start from the third level "###", while avoiding "#" and "##".

Please limit your response to {{ response_token_budget }} tokens.
\end{lstlisting}

\noindent User prompt:
\nopagebreak
\begin{lstlisting}[frame=tb]
--- Previous hints ---

{{ previous_hints }}

--- A new task-solving trajectory ---

{{ trajectory }}

Reward: {{ reward }}

Environment feedback: {{ feedback }}

--- Your job ---

Now, please think through it and return your updated hints based on the above information, following the requirements in the system prompt.
\end{lstlisting}
\bigskip

\begin{lstlisting}[frame=tb, caption={Prompts for solve-task episodes in \flobs. Note that the abstract symbols A/B/C/D used in the main text correspond to ``Direction 1'' through ``Direction 4'' in the prompt below.}, label={tab:draft_prompt_frozenlake_obscure_sys}]
You are playing a FrozenLake-Obscure game, where the action-to-direction mapping is unknown. Your task is to reach the goal (G) from your current position (P).

## Symbols
- P: Player (your current position)
- G: Goal (destination)
- _: Frozen tile (safe to walk)
- O: Hole (fall in and lose)

## Rules
1. Avoid falling into holes (O).
2. Reach the goal (G) to win.
3. The map is a grid with boundaries at its edges. If you try to move beyond the edge, your position will not change and the map will remain the same.

## Valid Actions
You can execute one of the following actions at each step: Direction 1, Direction 2, Direction 3, Direction 4.

## Hidden Clues About the Environment
Each valid action corresponds to moving along one of the up/down/left/right directions, but the exact action-to-direction mapping is unknown. Do not make any prior assumption about the mapping. You may try to figure it out through trial-and-error, e.g., by observing how your position (P) changes on the map after executing an action. You may also utilize additional hints provided in the context.

## Rewards
- Fall into hole (O): 0
- Reach goal (G): 1

## Response Format
First think about which action to take at the current step, then provide your final answer. Format your complete response as follows:
```
[THINKING]
<answer>Direction X</answer>
```
where [THINKING] should be replaced with your thinking process, and X is one of 1/2/3/4.

Other requirements:
- Be concise and avoid overthinking in your thinking process.
- Take one and only one action for each step.
- You should reach the goal (G) within a limited number of steps. Try to find a short path towards the goal.
- Balance exploration and exploitation: as you strive to reach the goal, you may acquire information about the environment along the way, which can be potentially helpful for solving other similar tasks later on.
{%

## Hints that might help
Below are some hints that might be helpful, but there is no guarantee that they must be correct or applicable to the current task. You might leverage them as prior knowledge, while incorporating new information from your own experience in interacting with the environment.

{{ hint }}
{%

## Response length limit
Please limit your response (including your thinking process) to {{ response_token_budget }} tokens.
\end{lstlisting}

\noindent Per-step user prompt:
\nopagebreak
\begin{lstlisting}[frame=tb]
Step {{ current_step }}/{{ max_steps }}
{%

{{ action_feedback }}
{%

Current map:
{{ observation }}
{%
You have not reached the goal (row={{ goal_row }}, col={{ goal_col }}) yet. Please decide your next action.
{%
\end{lstlisting}
\bigskip

\begin{lstlisting}[frame=tb, caption={Prompts for solve-task episodes in \alchrand.}, label={tab:draft_prompt_alchemy_random_sys}]
You are in an Alchemy crafting environment. Your goal is to synthesize the target element "{{ target }}" by combining elements in your inventory.

## Game Mechanics

There exists a fixed set of recipes. Each recipe takes exactly two input elements and produces one output element. A recipe's two inputs can be the same or different elements, and can be from different tiers.

- **Combination**: Each round, choose two elements from your inventory to combine. If the pair matches a valid recipe, the result is added to your inventory. If not, nothing is produced.
- **Tier system**: Every element has a tier. Every recipe produces a result whose tier is strictly higher than both inputs' tiers.
- **No consumption**: Elements are NOT consumed when used. You keep all elements after each attempt, whether it succeeds or fails. Newly discovered elements are permanently added to your inventory.
- **Winning**: You win when "{{ target }}" appears in your inventory.
- **Ending**: The game ends when you succeed or reach the round limit.
- All element names in this Alchemy crafting environment have no correspondence to real-world concepts. Do not try to guess recipes based on element names.

## Hidden Clues About the Environment
Recipes (i.e., which elements can be combined and which cannot) are unknown initially. You can try to discover them through experimentation, or based on additional hints provided in the context.

## Element Tiers

{{ tier_info }}

## Response Format
First think about which combination of elements to try out in the current round, then provide your action using <answer>...</answer> tags at the end. Format your complete response as follows:
```
[THINKING]
<answer>Element1 + Element2</answer>
```
where [THINKING] should be replaced with your thinking process, while Element1 and Element2 are elements (same or different) chosen from your inventory.

Other requirements:
- Be concise and avoid overthinking in your thinking process.
- Take one and only one action for each round.
- You should synthesize the target element within a limited number of rounds. Try to find a short path towards the goal, e.g., by leveraging recipes that have been revealed and avoiding combinations that are known to fail.
- Balance exploration and exploitation: as you strive to reach the goal, you may acquire information about the environment along the way, which can be potentially helpful for solving other similar tasks later on.
{%

## Hints that might help
Below are some hints that might be helpful, but there is no guarantee that they must be correct or applicable to the current task. You might leverage them as prior knowledge, while incorporating new information from your own experience in interacting with the environment.

{{ hint }}
{%

## Response length limit
Please limit your response (including your thinking process) to {{ response_token_budget }} tokens.
\end{lstlisting}

\noindent Per-round user prompt:
\nopagebreak
\begin{lstlisting}[frame=tb]
Round {{ current_round }}/{{ max_rounds }}
{%

{{ action_feedback }}
{%

Target: {{ target }}

Available elements in your inventory:
{%
- {{ elem }}
{%
{%

Discovered recipes:
{%
- {{ r[0] }} + {{ r[1] }} = {{ r[2] }}
{%
{%
{%

Failed combinations:
{%
- {{ f[0] }} + {{ f[1] }}
{%
{%

Choose two elements available in your inventory to combine.
\end{lstlisting}
\bigskip

\begin{lstlisting}[frame=tb, caption={Prompts for solve-task episodes in \terminal.}, label={tab:draft_prompt_terminal_sys}]
You are interacting with a simulated terminal environment. Your goal is to complete file operations by typing commands.

## Environment
- Local machine: {{ os_name }} (type: {{ os_type }}, hostname: localhost)
- Remote server: {{ remote_host }} (Linux, SSH pre-configured)
- Remote username: {{ remote_user }}

## Available Commands
ls, dir, cd, pwd, cat, type, mkdir, chmod, cp, copy, mv, move, rm, del, ren, touch, scp, rsync, ssh, exit, whoami, echo, tar, zip, unzip, gzip, gunzip

Note: Not all commands are available on all systems.

## Task
{{ task_description }}

## Response Format
First think step by step about which command to run, then provide your command using <answer>...</answer> tags at the end. Format your complete response as follows:
```
[THINKING]
<answer>COMMAND</answer>
```
where [THINKING] should be replaced with your thinking process, while COMMAND is a single shell command line to execute this turn.

Other requirements:
- Be concise and avoid overthinking in your thinking process.
- Issue exactly ONE command per turn. You may use `&&` or `;` to chain commands within the single `<answer>` block.
- If your response cannot be parsed (missing `<answer>` tags), the episode ends immediately.
- You have {{ max_steps }} steps to complete the task.
- The `<answer>` tag must appear at the END of your response.
{%

## Hints that might help
Below are some hints that might be helpful, but there is no guarantee that they must be correct or applicable to the current task. You might leverage them as prior knowledge, while incorporating new information from your own experience in interacting with the environment.

{{ hint }}
{%

## Response length limit
Please limit your response (including your thinking process) to {{ response_token_budget }} tokens.
\end{lstlisting}

\noindent Per-step user prompt:
\nopagebreak
\begin{lstlisting}[frame=tb]
Step {{ current_step }}/{{ max_steps }}
{%

{{ command_output }}
{%

{{ terminal_prompt }}
\end{lstlisting}

\section{Examples of hints}
\label{subsec:example_hint}

Tables~\ref{tab:hint_example_flobs}, \ref{tab:hint_example_alchemy} and \ref{tab:hint_example_terminal} present examples of hints generated by the agent in update-context episodes.

\begin{lstlisting}[frame=tb, caption={A hint generated for \textsc{FrozenLake-Obscure}, recovering the action-to-direction mapping.}, label={tab:hint_example_flobs}]
[THINKING]
The task involved navigating a 5x5 grid to reach the goal while avoiding holes. The action-to-direction mapping was initially unknown, but it was determined through trial and error. Direction 4 corresponds to left, Direction 1 corresponds to right, Direction 3 corresponds to down, and Direction 2 corresponds to up. The player started at (2,3) and moved left to (2,2), then right to (2,3), then down to (3,3), then right to (3,4), then down to (4,4). The key insights are the action-to-direction mapping and the strategy for navigating the grid. The updated hints should include the confirmed action-to-direction mapping, the strategy for navigating the grid, and the importance of checking the validity of each move.
--- Start of updated hints ---
### Action-to-direction mapping
- Direction 1 = right
- Direction 2 = up
- Direction 3 = down
- Direction 4 = left
### Navigation strategy
- Move in the direction that leads to the goal
- Avoid moving in the direction that leads to a hole
- Plan a safe path to the goal
- Use the action-to-direction mapping to navigate the map
### General advice
- Check the validity of each move before executing it
- Use trial and error to determine the action-to-direction mapping
- Balance exploration and exploitation to find the optimal path
--- End of updated hints ---
\end{lstlisting}
\bigskip

\begin{lstlisting}[frame=tb, caption={A hint generated for \alchrand, recording discovered crafting recipes.}, label={tab:hint_example_alchemy}]
[THINKING]
The task involved synthesizing the target element "fhsj" by following a specific recipe path. Based on the successful trajectory, we can derive several key insights. First, the target element "fhsj" is at tier 3, which means it can only be synthesized by combining two tier 2 elements. The only tier 2 element available is "joouk", so we need to synthesize "joouk" first. To synthesize "joouk", we need to combine two tier 1 elements. The tier 1 elements are "hjklt" and "nlji", so we need to synthesize both of these elements first. To synthesize "hjklt", we need to combine two tier 0 elements. The tier 0 elements are "lqlrr", "gwzzz", "bbtrt", and "vbpzh". Based on the successful trajectory, we found that "lqlrr + gwzzz" produces "hjklt", "lqlrr + vbpzh" produces "nlji", and "hjklt + nlji" produces "joouk". Therefore, the optimal path to synthesize "fhsj" is to first synthesize "hjklt" by combining "lqlrr" and "gwzzz", then synthesize "nlji" by combining "lqlrr" and "vbpzh", then synthesize "joouk" by combining "hjklt" and "nlji", and finally synthesize "fhsj" by combining "joouk" and "joouk".
--- Start of updated hints ---
### General Strategy
- Synthesize elements in ascending tier order to reach the target element
- Always prioritize synthesizing the elements required for the next tier
### Tier 0 to Tier 1
- **lqlrr + gwzzz** = hjklt
- **lqlrr + vbpzh** = nlji
### Tier 1 to Tier 2
- **hjklt + nlji** = joouk
### Tier 2 to Tier 3
- **joouk + joouk** = fhsj
### Key Recipes
- **lqlrr + gwzzz** = hjklt
- **lqlrr + vbpzh** = nlji
- **hjklt + nlji** = joouk
- **joouk + joouk** = fhsj
### Strategy for Synthesizing fhsj
1. Synthesize **hjklt** by combining **lqlrr** and **gwzzz**
2. Synthesize **nlji** by combining **lqlrr** and **vbpzh**
3. Synthesize **joouk** by combining **hjklt** and **nlji**
4. Synthesize **fhsj** by combining **joouk** and **joouk**
--- End of updated hints ---
\end{lstlisting}
\bigskip

\begin{lstlisting}[frame=tb, caption={A hint generated for \terminal, distilling reusable command sequences.}, label={tab:hint_example_terminal}]
[THINKING]
The provided task-solving trajectory demonstrates an effective approach to downloading and extracting a zip file from a remote server. The user used the `scp` command to copy the file from the remote server to their local machine and then used the `unzip` command to extract the contents to the specified directory. This approach is efficient and straightforward, and it can be applied to similar tasks. Based on this information, I will update the hints to include the following:

1. Use the `scp` command to copy files from a remote server to your local machine.
2. Use the `unzip` command to extract the contents of a zip file to a specified directory.
3. Navigate to the target directory before extracting the zip file using the `cd` command.

These hints are informative and can be helpful for future tasks that involve downloading and extracting files from a remote server.
--- Start of updated hints ---
- Use the `scp` command to copy files from a remote server to your local machine.
- Use the `unzip` command to extract the contents of a zip file to a specified directory.
- Navigate to the target directory before extracting the zip file using the `cd` command.
--- End of updated hints ---
\end{lstlisting}

\clearpage
\newpage

\section{Analysis: learning a context-aware policy model}
\label{sec:appendix_dynamics}

One high-level idea behind \codtrain is to learn policies that condition their actions on the current
environment context rather than rely on a context-independent prior or bias.  
We find that this relies critically on a balanced distribution of training tasks: for example, if a particular action is optimal too often during training, the policy model can increase its average reward by favoring that action regardless of context, which contradicts the objective of \codtrain.

In this section, we provide in-depth analysis of this effect in a minimal two-action contextual bandit setting trained with on-policy REINFORCE.  
We prove formally that on-policy REINFORCE converges to a benign context-aware policy if the distribution of training tasks is sufficiently balanced, whereas imbalance induces an action prior and poor out-of-distribution generalization for the learned policy.

\subsection{A minimal contextual bandit}
\label{subsec:dynamics_setup}

At the beginning of each episode, the environment samples a context
\begin{equation}
    \bz \in \left\{\bz^{(1)},\bz^{(2)}\right\},
    \qquad
    \bz^{(1)}=(0.6,0.4),\quad \bz^{(2)}=(0.4,0.6),
\end{equation}
with probabilities
$\Pr(\bz=\bz^{(1)})=\rho$ and
$\Pr(\bz=\bz^{(2)})=1-\rho$.  We call the training distribution perfectly balanced when
$\rho=1/2$, in which case the two contexts are equally likely.
The agent chooses $a\in\{1,2\}$ and receives the deterministic reward
$R(a,\bz)=z_a$.  Action 1 is optimal under $\bz^{(1)}$, whereas action 2 is
optimal under $\bz^{(2)}$.

Let $\sigma(x)\coloneqq\frac{1}{1+\exp(-x)}$.  
We consider a softmax policy $\pi_{\theta}$
with logits $w_1z_1+b_1$ and $w_2z_2+b_2$:
\begin{equation}
    \theta=(w_1,w_2,b_1,b_2), \quad
    \pi_\theta(a=1\mid \bz)
    =\sigma\big((w_1z_1 + b_1) - (w_2z_2 + b_2)\big)
    =\sigma\big(w_1z_1-w_2z_2+b_1-b_2\big).
    \label{eq:dynamics_policy}
\end{equation}
Note that the weight term $w_1z_1-w_2z_2$ varies with the context $\bz$, whereas the bias term $b_1 - b_2$ shifts the action-1
log-odds equally in both contexts and therefore represents a context-independent
action preference.  Define the
optimal-action margins as
\begin{align}
    m_1 = m_1(\theta) &\coloneqq 0.6w_1-0.4w_2+b_1-b_2, \\
    m_2 = m_2(\theta) &\coloneqq -0.4w_1+0.6w_2-b_1+b_2.
    \label{eq:dynamics_margins}
\end{align}
Here $m_1$ and $m_2$ are the logit advantages of the optimal actions under
$\bz^{(1)}$ and $\bz^{(2)}$, respectively.  The expected reward can be calculated by
\begin{equation}
    J_\rho(\theta)
    = 0.4+0.2\left[\rho\,\sigma(m_1)
    +(1-\rho)\sigma(m_2)\right].
    \label{eq:dynamics_objective}
\end{equation}

\subsection{Expected policy-gradient dynamics}
\label{subsec:dynamics_expected_gradient}

The infinite-sample limit of on-policy REINFORCE is the gradient flow
$\dot\theta=\nabla_\theta J_\rho(\theta)$.  Since an action-independent baseline
does not change this expected gradient, we use the zero baseline throughout.
Differentiating Eq.~\eqref{eq:dynamics_objective} gives
\begin{equation}
\begin{aligned}
    \dot w_1
    &=0.2\left[0.6\rho\,\sigma'(m_1)
      -0.4(1-\rho)\sigma'(m_2)\right],\\
    \dot w_2
    &=0.2\left[-0.4\rho\,\sigma'(m_1)
      +0.6(1-\rho)\sigma'(m_2)\right],\\
    \dot b_1&=-\dot b_2
    =0.2\left[\rho\,\sigma'(m_1)
      -(1-\rho)\sigma'(m_2)\right].
\end{aligned}
    \label{eq:dynamics_ode}
\end{equation}

\paragraph{Training with a balanced distribution.}

For $\rho=1/2$ and $\theta(0)=0$, symmetry in
Eq.~\eqref{eq:dynamics_ode} preserves $w_1=w_2$ and
$b_1=b_2=0$.  The dynamics reduce to
$\dot w_1=\dot w_2=0.02\,\sigma'(0.2w_1)>0$.  Hence both optimal-action
margins equal $0.2w_1\to+\infty$, and
$\pi_\theta(a=1\mid\bz^{(1)})=\pi_\theta(a=2\mid\bz^{(2)})\to1$.
More precisely, for gradient-flow time $t$,
$w_1(t)=w_2(t)=5\log(t/250)+o(1)$, while the biases remain zero.  The learned
decision rule therefore depends entirely on the context.

\paragraph{Transient collapse under imbalanced training.}

For $\rho>1/2$, Eq.~\eqref{eq:dynamics_ode} gives
$\dot b_1=-\dot b_2=0.05(2\rho-1)>0$ at initialization.  REINFORCE therefore
develops a preference for action~1, favored by the more frequent context.  It
can dominate the evidence in $\bz^{(2)}$ for a long training period, making
the policy select action~1 in both contexts despite the higher reward of a
context-dependent policy.

For $0<\rho<1$, this collapse is not the strict $t\to\infty$ limit.  Using
$\sigma'(m)\sim e^{-m}$ in Eq.~\eqref{eq:dynamics_ode}, the long-time
solution from zero initialization is
\begin{equation}
\begin{aligned}
    w_1(t)
    &=5\log(t/125)+2.6\log\rho+2.4\log(1-\rho)+o(1),\\
    w_2(t)
    &=5\log(t/125)+2.4\log\rho+2.6\log(1-\rho)+o(1),\\
    b_1(t)&=-b_2(t)
    \longrightarrow 0.2\log\frac{\rho}{1-\rho}.
\end{aligned}
    \label{eq:dynamics_imbalanced_asymptotic}
\end{equation}
Equivalently, the two optimal-action margins are asymptotic to
$\log(\rho t/125)$ and $\log((1-\rho)t/125)$; both eventually diverge to
$+\infty$, but the unequal biases remain.  Imbalance therefore leaves a
persistent action prior and can cause a long-lived finite-time collapse.  For
$\rho<1/2$, the actions in these results are exchanged.

\subsection{Learning dynamics under policy capacity constraint}
\label{subsec:dynamics_weight_decay}

The unregularized dynamics can eventually overcome the action prior because
$w_1$ and $w_2$ are free to grow without bound.  To study the effect of finite policy capacity, we use bounded parameter scale as a proxy and add an
$\ell_2$ penalty, equivalent to weight decay under gradient flow:
\begin{equation}
    J_{\lambda,\rho}(\theta)=J_\rho(\theta)-\frac{\lambda}{2}
    \left(w_1^2+w_2^2+b_1^2+b_2^2\right),
    \qquad \lambda>0.
    \label{eq:dynamics_regularized_objective}
\end{equation}
The regularized dynamics add
$-\lambda\theta$ to the gradient flow in Eq.~\eqref{eq:dynamics_ode}.

The probabilities $\sigma(m_1)$ and $\sigma(m_2)$ impose only two constraints
on four parameters, so many parameter vectors represent the same policy.  The
$\ell_2$ penalty selects the unique minimum-norm vector, in the coordinate
order $(w_1,w_2,b_1,b_2)$:
\begin{equation}
    \theta^\star(m_1,m_2)
    =\bigl(2.6m_1+2.4m_2,\ 2.4m_1+2.6m_2,\
           0.2(m_1-m_2),\ -0.2(m_1-m_2)\bigr).
    \label{eq:dynamics_min_norm_parameters}
\end{equation}
Thus $b_1^\star=-b_2^\star=0.2(m_1-m_2)$: equal margins give zero biases, while
$m_1>m_2$ shifts the action-1 log-odds upward in both contexts.  The weights may
still favor action~2 under $\bz^{(2)}$.

Substituting Eq.~\eqref{eq:dynamics_min_norm_parameters} into the
regularized objective gives
\begin{equation}
\begin{aligned}
    J_{\lambda,\rho}\bigl(\theta^\star(m_1,m_2)\bigr)
    ={}&0.4+0.2\left[\rho\,\sigma(m_1)
        +(1-\rho)\sigma(m_2)\right]\\
       &-\frac{25\lambda}{4}(m_1+m_2)^2
        -\frac{\lambda}{20}(m_1-m_2)^2.
\end{aligned}
    \label{eq:dynamics_margin_objective}
\end{equation}
This reduction makes the effect of the training distribution explicit.  When
$\rho=1/2$, Eq.~\eqref{eq:dynamics_margin_objective} is symmetric in
$m_1$ and $m_2$, and its optimum satisfies $m_1^\star=m_2^\star$.
Eq.~\eqref{eq:dynamics_min_norm_parameters} then gives
$b_1^\star=b_2^\star=0$.
For an imbalanced distribution, the directional derivative that increases the
more frequent context's margin and decreases the other is, at equal margins,
\begin{equation}
    \left.\frac{\diff}{\diff\epsilon}
    J_{\lambda,\rho}\bigl(\theta^\star(m+\epsilon,m-\epsilon)\bigr)
    \right|_{\epsilon=0}
    =0.2(2\rho-1)\sigma'(m).
    \label{eq:dynamics_imbalance_derivative}
\end{equation}
It is nonzero for $\rho\neq1/2$, so equal margins cannot be stationary.  For
$\rho>1/2$, the gradient favors a larger first-context margin and, by
Eq.~\eqref{eq:dynamics_min_norm_parameters}, an action-1 bias shared
across contexts.

\begin{figure}
    \centering
    \includegraphics[width=.95\linewidth]{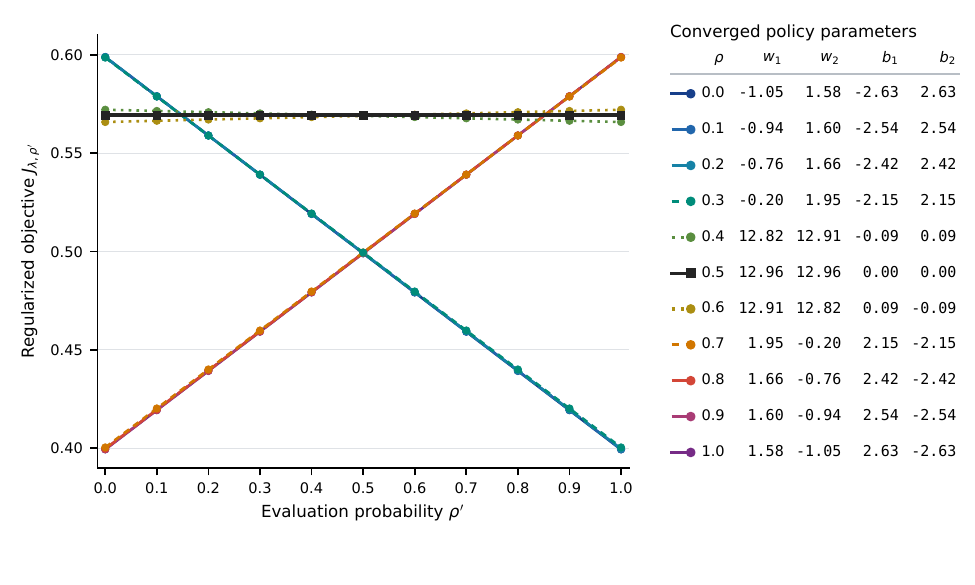}
    \vspace{-0.5em}
    \caption{Regularized objectives of 11 converged policies at
    $\lambda=10^{-4}$.  Each policy is obtained by exact expected REINFORCE
    gradient flow from $\theta(0)=0$ at a different training probability
    $\rho\in\{0,0.1,\ldots,1\}$, and its converged parameters are listed in the legend.
    Each curve then evaluates that fixed policy over $\rho'\in[0,1]$, with
    markers at all 11 labeled evaluation probabilities.  Curves within
    the groups $\rho\leq0.3$ and $\rho\geq0.7$ nearly overlap because their
    respective policies strongly favor the same action in both contexts.}
    \label{fig:dynamics_objective_tradeoff}
\end{figure}

For each $\rho\in[0,1]$, we integrate the regularized expected gradient flow
from $\theta(0)=0$ to a stationary policy $\theta_\rho$, with margins $m_1$
and $m_2$.  They satisfy
\begin{equation}
\begin{aligned}
    0.2\rho\sigma'(m_1)
    &=12.5\lambda(m_1+m_2)+0.1\lambda(m_1-m_2),\\
    0.2(1-\rho)\sigma'(m_2)
    &=12.5\lambda(m_1+m_2)-0.1\lambda(m_1-m_2).
\end{aligned}
    \label{eq:dynamics_stationary_system}
\end{equation}
Eq.~\eqref{eq:dynamics_min_norm_parameters} recovers $\theta_\rho$
from these margins.  The stationary equations can admit more than one solution;
the zero-initialized gradient flow selects the one plotted in Figure~\ref{fig:dynamics_objective_tradeoff}.

To assess the sensitivity of a particular policy $\theta_{\rho}$ --- trained under a task distribution specified by the probability $\rho$ --- we evaluate its objective value under a different task distribution specified by $\rho'$.
Training under a balanced distribution with $\rho = 1/2$ gives equal margins, so
\begin{equation}
    J_{\lambda,\rho'}(\theta_{1/2})
    =J_{\lambda,1/2}(\theta_{1/2})
    \quad\text{for all }\rho'.
    \label{eq:dynamics_uniform_transfer}
\end{equation}
For any fixed $\theta_\rho$,
\begin{equation}
    \frac{\partial}{\partial\rho'}
    J_{\lambda,\rho'}(\theta_\rho)
    =0.2\left[\sigma(m_1)-\sigma(m_2)\right],
    \label{eq:dynamics_distribution_sensitivity}
\end{equation}
which vanishes for the balanced policy and changes sign with its relative
success in the two contexts.

Figure~\ref{fig:dynamics_objective_tradeoff} evaluates 11 policies trained
from zero initialization at $\rho=0,0.1,\ldots,1$.  Each line evaluates one
fixed policy as $\rho'$ varies from zero to one.  These lines are affine by
Eq.~\eqref{eq:dynamics_distribution_sensitivity}; their markers show
the 11 labeled evaluation mixtures, including each policy's training
mixture $\rho'=\rho$.  The balanced policy remains at $0.5693$ for every
$\rho'$.  At $\rho=0.4$ or $0.6$, the corresponding curves remain nearly flat; each policy's in-distribution objective $J_{\lambda,\rho}(\theta_\rho)$ exceeds $J_{\lambda,\rho}(\theta_{1/2})$ by only about $3\times10^{-4}$.  The policies
trained at $\rho\leq0.3$ or $\rho\geq0.7$ instead have large action biases (due to premature convergence to undesirable local optima)
and are much more sensitive to the evaluation mixture parameter $\rho'$.

\subsection{Finite-sample verification of the expected dynamics}
\label{subsec:dynamics_finite_sample}

We conduct numerical experiments with finite-sample gradient updates.
At update $k$, we independently sample a batch of $B$ contexts $\bz_{k,j}$
and actions $a_{k,j}\sim\pi_{\theta_k}(\cdot\mid\bz_{k,j})$.  With learning
rate $\eta$, REINFORCE uses their mean gradient:
\begin{equation}
    \theta_{k+1}=(1-\eta\lambda)\theta_k
    +\frac{\eta}{B}\sum_{j=1}^{B}R(a_{k,j},\bz_{k,j})\nabla_\theta
    \log\pi_{\theta_k}(a_{k,j}\mid \bz_{k,j}).
    \label{eq:dynamics_stochastic_update}
\end{equation}
Its expected direction is Eq.~\eqref{eq:dynamics_ode} plus the
$-\lambda\theta$ decay term; sampling introduces zero-mean noise around this update but does not change its expectation conditional on $\theta_k$.

\begin{figure}
    \centering
    \includegraphics[width=\linewidth]{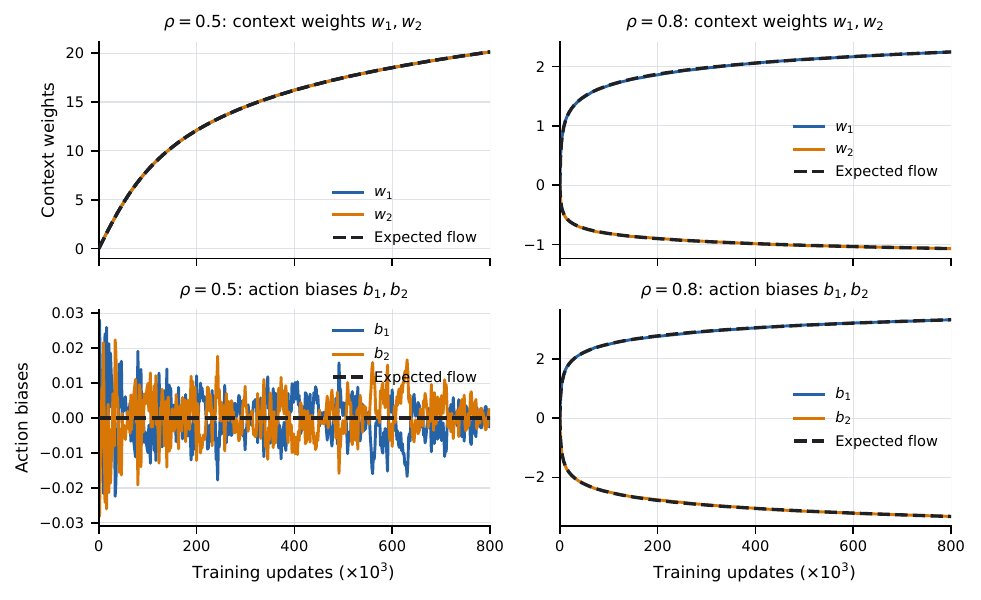}
    \caption{Expected (black dashed) and finite-sample (colored solid)
    REINFORCE dynamics without weight decay from $\theta(0)=0$.  Training uses
    $\eta=0.02$, batch size $B=128$, and $800{,}000$ updates.  Balanced sampling yields equal
    expected weights and zero expected biases; imbalanced sampling develops a
    majority-action preference over the plotted horizon.}
    \label{fig:dynamics_parameter_trajectories}
\end{figure}

Figure~\ref{fig:dynamics_parameter_trajectories} compares a representative
sampled trajectory with the expected gradient flow in the unregularized case.
At $\rho=0.5$, the weights grow together and the biases remain near zero.  At
$\rho=0.8$, the biases favor action~1 strongly enough to override the less
frequent context over the plotted horizon.  The sampled trajectory follows the
expected flow with stochastic fluctuations.  Without weight decay, the
balanced weights have no finite stationary limit.

\begin{figure}
    \centering
    \includegraphics[width=\linewidth]{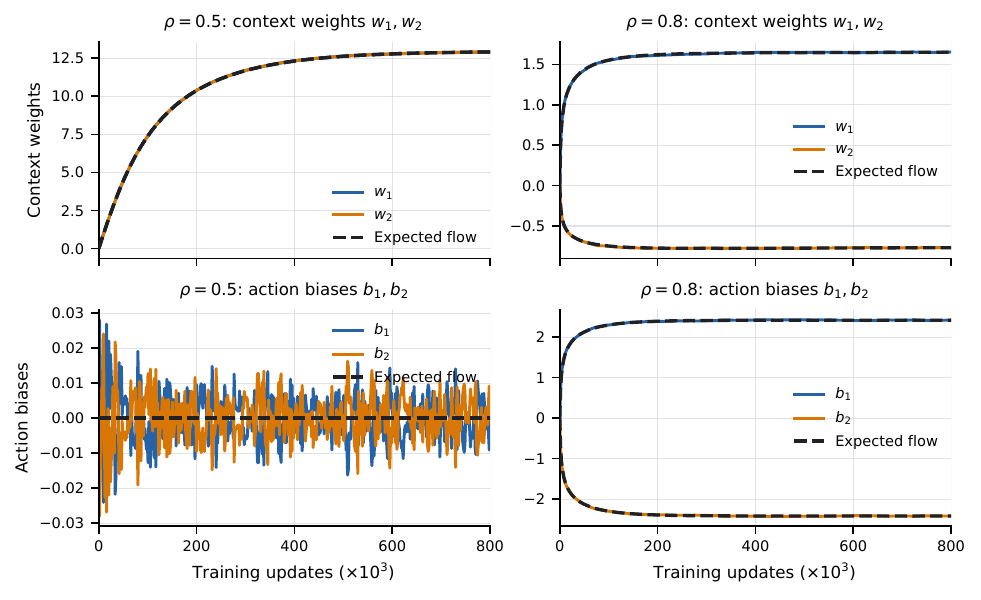}
    \caption{Expected (black dashed) and finite-sample (colored solid)
    REINFORCE dynamics with weight decay $\lambda=10^{-4}$ from
    $\theta(0)=0$.  Learning rate $\eta=0.02$, batch size $B=128$, context
    probabilities, and the $800{,}000$-update horizon match
    Figure~\ref{fig:dynamics_parameter_trajectories}.  The expected flow
    approaches stationarity, and the sampled trajectory fluctuates around it.
    Balanced sampling keeps the expected biases at zero; imbalanced sampling
    retains a majority-action preference.}
    \label{fig:dynamics_parameter_trajectories_weight_decay}
\end{figure}

Figure~\ref{fig:dynamics_parameter_trajectories_weight_decay} repeats the
experiment with $\lambda=10^{-4}$ over the same training horizon.  The
expected parameters approach finite stationary values, and the imbalanced
trajectory retains its majority-action preference.  Under balanced sampling,
the biases fluctuate around zero.

\end{document}